\documentclass{article}
\usepackage{arxiv}

\usepackage{xcolor}
\usepackage{hyperref}
\usepackage{graphicx}
\usepackage{enumitem}
\usepackage{bbm}
\usepackage{multirow}
\usepackage{caption}
\usepackage{subcaption}
\usepackage{amsmath,amsfonts,amssymb,amsthm}
\usepackage{xcolor}
\usepackage{natbib}
\usepackage{url}

\newcommand{\indep}{\perp \!\!\! \perp}

\newcommand{\M}{\mathcal{M}}
\newcommand{\R}{\mathcal{R}}

\newcommand{\E}{\mathbb{E}}
\newcommand{\Ex}[1]{\mathbb{E}\left[#1\right]}
\newcommand{\var}[1]{\mathrm{Var}\left(#1\right)}
\newcommand{\Cov}{\text{Cov}}
\newcommand{\Var}{\text{Var}}

\newcommand{\bz}{\boldsymbol{Z}}
\newcommand{\bc}{\boldsymbol{\mathcal{C}}}

\newcommand{\bA}{\boldsymbol{\mathcal{A}}}
\newcommand{\bm}{\boldsymbol{\mathcal{\mu}}}
\newcommand{\bU}{\boldsymbol{U}}

\newcommand{\am}{\boldsymbol{\mathcal{A}}\boldsymbol{\mu}}
\newcommand{\newG}{P_{\bz}(\bm)}
\newcommand{\pr}[1]{P\left(#1\right)}
\newcommand{\lo}{\text{L}}
\newcommand{\se}{\text{SE}}

\newtheorem{theorem}{\bf Theorem}[section]

\newtheorem{corollary}{\bf Corollary}[section]
\newtheorem{lemma}{\bf Lemma}[section]

\title{On the meaning of uncertainty for ethical AI: philosophy and practice}

\author{ \href{https://orcid.org/0009-0004-5968-2496}{\includegraphics[scale=0.06]{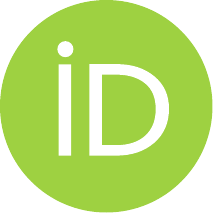}\hspace{1mm}Cassandra Bird}\thanks{Corresponding author.} \\
	Department of Mathematics and Statistics\\
	University of Exeter\\
	\texttt{crb226@exeter.ac.uk} \\
	\And
	\href{https://orcid.org/0000-0001-8917-3300}{\includegraphics[scale=0.06]{orcid.pdf}\hspace{1mm}Daniel Williamson} \\
	Land, Environment, Economics and Policy Institute\\
	University of Exeter\\
	\\
	\AND
	\href{https://orcid.org/0000-0002-7815-6609}{\includegraphics[scale=0.06]{orcid.pdf}\hspace{1mm}Sabina Leonelli} \\
	Exeter Centre for the Study of the Life Sciences (Egenis)\\
	University of Exeter\\
}

\begin{document}
\maketitle


\begin{abstract}
Whether and how data scientists, statisticians and modellers should be accountable for the AI systems they develop remains a controversial and highly debated topic, especially given the complexity of AI systems and the difficulties in comparing and synthesising competing claims arising from their deployment for data analysis. This paper proposes to address this issue by decreasing the opacity and heightening the accountability of decision making using AI systems, through the explicit acknowledgement of the statistical foundations that underpin their development and the ways in which these dictate how their results should be interpreted and acted upon by users. In turn, this enhances (1) the responsiveness of the models to feedback, (2) the quality and meaning of uncertainty on their outputs and (3) their transparency to evaluation. To exemplify this approach, we extend Posterior Belief Assessment to offer a route to belief ownership from complex and competing AI structures. We argue that this is a significant way to bring ethical considerations into mathematical reasoning, and to implement ethical AI in statistical practice. We demonstrate these ideas within the context of competing models used to advise the UK government on the spread of the Omicron variant of COVID-19 during December 2021. 
\end{abstract}

\section{Introduction}\label{sec::intro}

With structures that typically consist of thousands of layers and parameters, large statistical models such as Deep Gaussian Processes \citep{DGP} and Neural Networks (for example, see \cite{Modern}) fall well within the scope of the definition of Artificial Intelligence (AI), even as this domain as a whole remains hard to define.\footnotemark[1] These large statistical models or `AI systems' are often being deployed, or considered for deployment, in critical decision-making scenarios. The dangers of such deployment have been highlighted in areas such as medical diagnoses \citep{MITAI} and facial recognition \cite{GS1}.

\footnotetext[1]{\cite{EURO} highlight the recent struggle in deciding how it should be established in the AI Act \citep{AIAct}, with the proposal maintaining that the term AI encompasses a diverse range of methods. These include (rather broadly) \textit{statistical approaches} and \textit{Bayesian estimation}.}

With the rise of AI has come the pressure to define and enforce ethical principles to regulate its use, and yet, it is no easier to define AI ethics than it is to define AI itself (see \cite{saltz2019data} for a recent systematic literature review conducted with the aim of establishing this definition in relation to data science). In discussing the effectiveness of ethical codes of conduct in this setting, \cite{whittaker2018ai} acknowledge the risk that they `deflect criticism by acknowledging that problems exist, without ceding any power to regulate or transform the way technology is developed and applied'. To make things worse, attempts at legislation and proposed governance frameworks \citep{AIA} are often predominantly data-focused, committing to broad objectives for vaguely defined systems when addressing the underlying mathematics. 

Various authors (including \cite{engagingtheethics}, \cite{leonelli2021data}, etc.) address this vagueness and the subsequent disconnect between data scientists and the actions they feel are available to them to address ethical issues. They highlight the need for techniques that help to bridge the gap between these disciplines and related concerns, and the importance of thinking about modelling strategies in particular as crucial in this respect. \cite{saltz2019data} also acknowledge concerns over modelling choices, and many philosophers have pointed to the role that assumptions and mathematical tools can play in addressing ethical issues. That being said, we find that there exists few spaces in which issues that are advanced in a technical sense, whilst also being significant from an ethical perspective, can be effectively tackled in an interdisciplinary manner (whether this be due to reluctance on either side to engage with unfamiliar theory or a lack of a shared vocabulary). In this paper, we aim to circumvent this issue by highlighting our ethical concerns regarding how statistical theory is used in AI systems such that it is accessible to both sides respectively. 

Acknowledging modelling choices and their underpinning foundations should play a key role in shaping the outputs of AI and the meaning that can subsequently be attributed to those outputs. This, in turn, dictates how the outputs of AI can or should be used to support decision making. The statistical foundations refer to the meaning of uncertainty, directly establishing how it can be quantified and used in practice. 

In their account of decision theory, \cite{Jim} explains that the justification of one's decisions relies on one's ownership of the uncertainty utilised in the decision-making process, with all probabilities required to have come from `you or an expert you trust'. A common understanding, between modeller and decision maker, of the meaning of probabilities used for decision making (whether the probabilities are made explicit to the decision maker or are used internally by the AI system) is critical for justifiable decision making. This implies that the modeller themselves, whatever the scenario, must understand and be able to communicate the foundational implications of their modelling judgements.

Consider, for example, a Neural Network (NN) classifier with a fixed structure of hidden layers and transfer functions. If the network represents an explicit statistical model with a multinomial output layer, the parameters can either be fitted by maximum likelihood or given a prior; turning the model into a Bayesian Neural Network that can be fitted directly via variational approximation or using an approximate Gaussian process representation. In both cases the AI describes the classification problem with the same probability model, yet the probabilities are foundationally different. Whilst it's impossible to directly ascribe precise meaning to either set of probabilities without knowing the context in which the modeller intended them, broadly, the former represent (estimates of) `generative' probabilities that describe the relative frequency of the different classes amongst individuals with identical features, and the latter represent the subjective uncertainty of the modeller. Even here, whether the subjective uncertainty applies to the generative probabilities themselves \citep{Berger}, or exists to express uncertainty about an unclassified individual without admitting the existence of generative probabilities \citep{Goldstein}, depends on the intent of the modeller.

The use of an explicit probability model at least gives a starting point for a conversation about meaning. The fixed structure NN in our example above might instead be fitted by optimising a score based on correctly partitioning a training set, avoiding an explicit modelling step. However, the foundations are still at work. Here, to use the `optimal' NN out of sample both implies a specific value judgement made by the modeller that is inherent in whatever score is used, and a judgement of exchangeability of the training set with a population (note that this judgement induces a probability model via the representation theorem \citep{Finetti}). 

In all cases, what it means to use an AI system is derived from the judgements and intentions of the experts whom we call `the modellers'. Within this context, who the modeller(s) is/are ranges from the data scientist fitting data on their laptop, to the developer(s) of broad-use black box implementations such as general-purpose AI \citep{generalpurpose}. In this paper, we argue that the ways in which these judgements and intentions are documented as well as communicated to users naturally lends itself to a discussion of ethics. In particular, we argue that when we consider what makes the use of AI ethical, an important component must be the acknowledgment that action informed by the uncertainty produced by a modeller (via an AI system), requires the adoption of that uncertainty by the user. For this to be a justifiable exchange, the modeller themselves must understand the foundational implications of their modelling judgements and be able to clearly communicate them to a user, thus enabling them to make the decision as to whether this is uncertainty they wish to act on. For example, is the modeller expressing their own uncertainty through probability and, if so, does the user understand that or do they think the modeller has measured a property of the world (moreover does that property of the world exist and what does it mean)? 

Unfortunately, in most cases regarding AI, these foundational judgements and intentions are not well understood by the modellers themselves, much less by any downstream users or decision makers. Even if a modeller and user understand the same foundational meaning for probabilities and judgements, can any modeller, let alone any user, `own' all of the complex judgements and statements made by an AI system? 
 
The fact that it is currently unclear what the answer to this question is in the case of a single modeller advising a single user through the use of one AI model is a cause for concern, particularly given the more realistic case in which multiple modellers will be advising multiple users through the fitting of a range of potentially contradictory systems. And yet, concepts such as general-purpose AI continue to actively exacerbate the ever-widening divide between modeller and user. Whilst we hold that shared understanding and ownership of foundational judgements should be a bare minimum for ethical use of AI - particularly in decision-making contexts - it is not possible to cover all possible worldviews and nuances these foundations might take and to do them justice, without misrepresentation, in a single paper. Instead we proceed from our own foundational position, the subjectivist position outlined by \cite{Finetti}: that `true randomness' need not exist and that probability is a measure of individual belief. From here we can examine belief ownership methodologies for complex AI systems and address this particular facet of the ethical use of AI. Whilst we do not wish to rehash old debates around competing foundational worldviews \citep[see][for a thorough account]{Gelman}, we do insist that the work of imbuing meaning and ownership for AI is required, whatever the worldview.  

Subjectivism and Bayesian inference are natural partners. The subjective Bayesian account holding that if the statistical model and the prior represent your initial beliefs before seeing the data, your beliefs are represented by the posterior distribution having seen it. In fact, the traditional Bayesian approach represents an impossibly high bar for belief ownership, particularly for complex AI systems.  \cite{danaher2016threat} defines opacity as the issue of AI systems working \textit{`in ways that are inaccessible or opaque to human reasoning and understanding'}\footnotemark[2]. From a foundational point of view, this opacity renders the traditional subjective Bayesian account of inference impossible from a practical perspective and calls into question the way in which (and the extent to which) responsibility and accountability should be related to the modellers. This has motivated other routes to subjective inference when utilising complex models. 

Common to all routes (that we are aware of) is the acknowledgement that the Data Generating Process (DPG) may not be recovered at any setting of the parameters of the model (we defer subjectivist interpretations of the DGP to later). \cite{Bernardo} termed this the `$\mathcal{M}$-open' view (where $\mathcal{M}$ indicates the class of models obtained by varying the parameters only), as opposed to the more traditional `$\mathcal{M}$-closed' view. Inference in the $\mathcal{M}$-open world has received a lot of attention recently in the context of highly complex statistical models such as Neural Networks, with two broad approaches being popular. Model Synthesis approaches seek to combine perturbations to the model class, or perhaps even competing models from different groups, in order to develop $\mathcal{M}$-open inference and prediction. Generalised Bayesian Inference \citep{Bissiri} seeks to efficiently provide uncertainty quantification for AI systems by treating Bayes as an optimisation problem. Specifically, posterior uncertainty is derived by optimising a loss function that includes a divergence between the model class and the DGP. We review these approaches in Section \ref{sec::m-open}. 

In this paper we examine an existing synthesis method, Posterior Belief Assessment (PBA, \cite{PBA}), as a means to belief ownership for decision-critical statements made by AI systems. We extend the approach for this purpose and advocate for it as one route towards these important considerations for the ethical use of AI. Within more traditional areas of ethical AI, we argue that the approach decreases the opacity and increases the accountability of predictive AI systems by addressing (1) the responsiveness of the models to feedback, by identifying moments of constructive dialogue between modellers and relevant users around the assumptions inserted into the model; (2) the explainability of the model outputs, by clarifying some of the key decisions and judgements incorporated into the model and their effects on the modeller's uncertainty regarding the analysis; and (3) the amenability of the models to evaluation, which is made easier by a stronger awareness of the assumptions (empirical and conceptual) employed in various stages of modelling, as well as who may be responsible for such assumptions. Notably, the strategy presented here is inspired as much from our statistical work as it is from our concerns with ethics: in our experience, the two can be successfully combined to yield advances in statistical modelling.

We present and extend PBA for ethical AI in Sections \ref{sec::pba1} and \ref{sec::pba2}. Section \ref{sec::casestudy} presents a detailed illustration of the practical ideas developed by application of the methodology to the use of competing models of the spread of the Omicron variant of COVID-19 for advising the UK government during the pandemic. Section \ref{sec::ethical.discussion} offers further discussion of the implications of the foundational considerations discussed in this paper to ethical AI, and our conclusions are presented in Section \ref{Future}.

\footnotetext[2]{\cite{danaher2016threat} reserves the term \textit{hiddenness} for data-related concerns, such as privacy.} 

\section{AI and the $\mathcal{M}$-open worldview}\label{sec::m-open}

$\mathcal{M}$-open approaches acknowledge that any AI system or `model' can only be an approximation to reality. We review the main $\mathcal{M}$-open approaches that attempt to quantify uncertainty in reality through complex AI systems.

\subsection{Generalised Bayesian Inference (GBI)}

Generalised Bayesian Inference (GBI) \citep{Bissiri} allows for uncertainty quantification, even for parameter dense statistical models such as Neural Networks, by recasting Bayesian updating as an optimisation problem of the form
\begin{equation}\label{GBI}
    \textit{q}^{*}(\boldsymbol{\theta}) = \underset{q\in\mathcal{Q}}{\text{argmin}} \Bigg\{\mathbb{E}_{\textit{q}(\boldsymbol{\theta})} \Bigg[\sum_{i=1}^{n}\ell(\boldsymbol{\theta},x_i)\Bigg] + \text{KLD}(\textit{q}\parallel\pi)\Bigg\}, 
\end{equation}
where $\textit{q}^{*}(\boldsymbol{\theta})$ is the posterior distribution, $\mathcal{Q}$ is the space of all possible distributions indexed by $\boldsymbol{\theta}$, $\ell(\boldsymbol{\theta},x_i)$ is an additive loss over the parameters $\boldsymbol{\theta}$ and data $x_i$, $\pi(\boldsymbol{\theta})$ is the prior and $\text{KLD}(\textit{q}\parallel\pi)$ is the Kullback-Leibler divergence, so that $$\text{KLD}(q||\pi) = \E_{q(\boldsymbol{\theta})}[\text{log}(q(\boldsymbol{\theta}))] - \E_{q(\boldsymbol{\theta})}[\text{log}(\pi(\boldsymbol{\theta}))].$$ 

The optimal probability measure, $q^*$, coincides with the traditional Bayesian posterior when $\ell(\boldsymbol{\theta},x_i)$ is taken to be the negative log-likelihood with respect to the model $p(x|\boldsymbol\theta)$, and this is the only solution under  $\mathcal{M}$-closed (by coherence). However, for $\mathcal{M}$-open inference, setting $\ell$ to be the negative log-likelihood defines $\boldsymbol\theta$ to be the parameter values that minimise the KLD between the model, $p(\boldsymbol{x}|\boldsymbol{\theta})$, and the DGP, $p_t(\boldsymbol{x})$: $$\text{KLD}(p_t(\boldsymbol{x})||p(\boldsymbol{x}|\boldsymbol{\theta})) = \E_{p_t(\boldsymbol{x})}[\text{log}(p_t(\boldsymbol{x})] - \E_{p_t(\boldsymbol{x})}[\text{log}(p(\boldsymbol{x};\boldsymbol{\theta}))].$$

\noindent Minimising this expression with respect to $\boldsymbol{\theta}$ amounts to minimising $- \E_{p_t(\boldsymbol{x})}[\text{log}(p(\boldsymbol{x};\boldsymbol{\theta}))],$ for which the negative log likelihood is used as a finite sample estimate. The expected negative log-likelihood, taken with respect to $q(\boldsymbol{\theta})$, forms the first term of the objective in \eqref{GBI}, corresponding to an alternative but equivalent formulation of Bayes' rule. 

The $\M$-open interpretation allows the foundational position of analyst and user to be expressed and understood, giving an explicit meaning to the parameters and model which can be carried through to predictive inference. For example, belief in the existence of a DGP, as though we are observing a random process and attempting to model/infer and predict it, can be easily interpreted from this position. The subjectivist view holds that the DGP can be considered to be a user's true beliefs/uncertainties regarding the data, with the model as an imperfect representation of true belief \citep{Jack2}. Bayes then minimises the distance between these true and approximated beliefs given the data.

\noindent Any proper scoring rule, $S$, gives rise to a divergence between distributions $g$ and $f$ via  
\begin{equation}\label{div}
\text{D}(g||f) = \E_{g}[S(x,g)] - \E_{g}[S(x,f)]   
\end{equation}
(for observations $x$). We can therefore consider particular types of divergence to be associated with particular proper scores. In particular, the KLD is associated with the log score. The KLD is sensitive to similarities/differences in the tails of distributions which, given the acknowledgement of model misspecification, may be inappropriate. \cite{Bissiri} propose alternative choices of scoring rule (through loss $\ell$ in \eqref{GBI}), resulting in `Generalised' Bayes that targets different divergences. \cite{Jack2} review divergences that have desirable qualities, such as the $\beta$-divergence (\cite{knoblauch2018doubly}, \cite{jewson2023stability}): $$D_{\beta}(p_t(x)||p(x|\boldsymbol{\theta})) = \frac{1}{\beta(\beta-1)}\int p_t(x)^{\beta}dx + \frac{1}{\beta}\int p(x|\boldsymbol{\theta})^{\beta}dx - \frac{1}{\beta-1}\int p_t(x)p(x|\boldsymbol{\theta})^{\beta-1} dx.$$ The minimisation of this expression corresponds to the choice of $\ell$ in \eqref{GBI} as $$\ell(\boldsymbol{\theta},x_i) = -\frac{1}{\beta-1}p(x_i|\boldsymbol{\theta})^{\beta-1} + \frac{1}{\beta}\int p(x_i|\boldsymbol{\theta})^{\beta}dx \qquad \beta\in\mathbb{R}\backslash\{0,1\}.$$ 
For $\beta>1$, the greater the value of $\beta$, the more robust the divergence is to data outliers (with respect to the model). 

The second term of \eqref{GBI}, $\text{KLD}(\textit{q}\parallel\pi)$, is a penalty induced by dissimilarity of the prior and posterior. Note that \eqref{GBI} amounts to minimising expected loss where the loss function is additive in the data and prior, trading off divergence between prior and posterior with the representation of the DGP through the model. \cite{knoblauch2022optimization} builds on the ideas presented by \cite{Bissiri} through experimenting with also altering the prior divergence, generalising $\text{KLD}(\textit{q}\parallel\pi)$ to more robust choices, $D(q||\pi)$ in \eqref{GBI} \citep{knoblauch2022optimization, jaskari2022uncertainty}. \cite{knoblauch2022optimization} also propose Generalised Variational Inference, restricting the optimisation over $q$ to a class of tractable distributions, $\Pi$, for computational efficiency.

The framing of Bayesian inference as an optimisation problem has seen renewed interest in its use for the types of large statistical models used in AI, with literature examples of extensions and application to deep Gaussian Processes and deep Neural Networks \citep{knoblauch2022optimization, husain2022adversarial, wild2022generalized}. GVI has also been extended to work with intractable likelihood models \citep{pacchiardi2021generalized} and doubly-intractable likelihoods in the continuous \citep{matsubara2022robust} and discrete \citep{matsubara2022generalised} cases. Foundationally, from the subjectivist worldview, whilst having the flexibility to express nuanced confidence in a model's capacity to capture uncertainty for a DGP is useful, specifying these beliefs through quantities such as divergences and their hyperparameters is as or more challenging than trying to do so through probability distributions. It is widely acknowledged that further work is needed to determine how divergences and/or scoring rules should be selected, how they can be communicated to a party that does not necessarily have a mathematical background and how the effects of such choices propagate through the systems \citep{Jack2, knoblauch2022optimization, jewson2023stability}.

\subsection{Model Synthesis}  

Model synthesis addresses the $\mathcal{M}$-open problem by considering a range of possible statistical models rather than just one, acknowledging that, particularly when AI systems are highly complex, an array of options might adequately represent a modeller's beliefs and yet produce different results. In particular, model synthesis is often presented as a means by which modellers can facilitate decision making in a range of high-stakes scenarios, such as those that arise within climate science \citep[e.g.][]{knutti2017climate, rougier2013second} and epidemiology \citep[e.g.][]{silk2022uncertainty, maishman2022statistical}.

Perhaps the most popular collection of ideas, particularly in climate science where the outputs of different climate models might be seen as equivalent to the predictions/inferences given by complex statistical models, involve \textit{model weighting}. I.e. inference or prediction is given via a weighted sum of the individual model inferences/predictions where the weights are either assumed equal or assigned using some measure of in-sample skill or performance \citep{knutti2017climate, wang2016multi}. Whilst all methods of model synthesis we are aware of amount, mathematically, to weighting individual models, it is important to distinguish the methods detailed below, where weights arise naturally from foundationally explicit statistical models, from more ad-hoc methodology that leads to particular model combinations because they have performed well, according to some user-defined criteria, on a training set. 

Consider a finite model set $M_1, \ldots, M_n$, a quantity of interest (QoI), $X$, and data, $D$. The posterior predictive density obtained via 
\begin{equation}\label{weighting}
p(X|D)=\sum_{i=1}^{n}\text{w}_i p(X|D,M_i),
\end{equation}
where $p(X|D,M_i)$ are the predictive densities under each $M_i$ and the $\text{w}_i\in[0,1]$ are model weights, represents a mixture-model representation for the quantity of interest. From a subjectivist point of view, the $\text{w}_i$ could always be determined by expert judgement, though this precludes the data having any influence on the final synthesis. 

Note that, if the densities $p(M_i|D)$ exist, for example, if meaningful prior judgements $p(M_i)$ can be specified, then $\text{w}_i=p(M_i|D)$ by coherence. Note further that if the model set is assumed to be a random sample from an, in principle, infinite model collection with measure $p(M|D)$, \eqref{weighting} is a Monte Carlo approximation to the true posterior predictive density. Bayesian Model Averaging \citep[BMA,][]{hoeting1999bayesian} specifies $p(M_i)$ and weights as described above. BMA converges asymptotically to the single model, $M_i$, which is closest in KLD to the DGP, $p_t(X|D)$. 
 
An alternative to BMA is stacking \citep{yao2018using}, which derives the weights by minimising a divergence between the model combination and the DGP:
\begin{equation*}
    \boldsymbol{\text{w}}^{*} = \underset{\boldsymbol{\text{w}}}{\text{argmin}}\;\;\text{D}\left(\sum_{i=1}^{n}\text{w}_ip(X|D,M_i)\;||\;p_t(X|D)\right).
\end{equation*}
Similar to GBI, by making the choice of divergence explicit, stacking approaches allow the user to define the desired `closeness' between the posterior predictive and the DGP through the divergence, yet with the same caveats over the difficulty in making these judgements in practice. 
Stacking was used during the COVID-19 pandemic to weight models from UK modelling groups in supporting government decision making \citep{SAGE2, SAGE3} and further explored and extended within this context by \cite{silk2022uncertainty}. It has since been applied to aiding in the sampling of multimodal Bayesian posterior distributions \citep{yao2022stacking} and has been extended to \textit{Bayesian hierarchical stacking} \citep{yao2022bayesian}.

\subsection{Posterior Belief Assessment}\label{sec::pba1}

The approaches discussed thus far fit and/or combine probability distributions in order to arrive at a single probability distribution for the QoI. \cite{PBA} argued that the burden placed on a user for owning uncertainty in the form of full probability distributions was too great and argued instead for a synthesis approach, Posterior Belief Assessment (PBA), that retained the complexity and computational intricacy of the complex statistical models, yet only required a limited form of belief ownership from the modeller for a small number of key observables.

The concept is that, whilst one can obtain $p(X \mid D, M_i)$ for $i = 1, \ldots, n$, one cannot and is not usually required to believe all statements made by a synthesis distribution $p(X \mid D)$, but rather a handful of expectations of functions of $X$ that may be critical to decision making. For example, $p(X>C\mid D)$ the probability of $X$ exceeding some critical threshold, $C$, (taken as the expectation for the indicator function of the exceedance event) may be required, or an expected utility $\E\left[U(X)\mid D\right]$. Instead of calculating these from the synthesis distribution, one can form explicit judgements about the decision critical quantities of interest and then update them with their expectations as derived under each of the different models. Underpinning this concept is the notion that expectation, and not probability, is the primitive quantity for measuring belief. Expectation as primitive underpins the subjectivist account of probability theory \citep{Finetti}, allowing expectations to be specified and owned, without existence of an underpinning probability distribution. In this account, the probability of an event, when it exists, is nothing more than the expectation of the indicator function for the event and probability distributions are the collection of all such expectations (in such cases, expectations as calculated by integrating a function over a probability distribution must coincide with those specified, by coherence). In the following formulation, we use $P(\cdot)$ to denote a primitive expectation (an expectation that is actually believed) and reserve $\E(\cdot)$ for expectations derived from probabilistic models, noting that we might still believe these expectations, if we believe the underlying model. 

Through access to $p(X \mid D, M_i)$ we can calculate $\bz = \{\Ex{X\mid D, M_i}$;  $i= 1, \ldots, n\}$, the set of posterior expectations for the QoI for each of the $n$ models under consideration. A PBA forms a dual Hilbert space over $X$ and $\bz$ with inner product $<X, Y> = P(X^TY)$, and uses orthogonal projection of beliefs onto linear combinations of $\bz$ in order to \textit{adjust} beliefs in the light of data. This projection gives $P_{\bz}\left[X\right]$, termed the \textit{posterior belief assessment} (PBA), as the value minimising $||X-P_{\bz}(X)||^2$ over all affine combinations of $\bz$. In fact the PBA is a specific Bayes Linear update (see Section 3.1 of \cite{BL}), with \begin{equation}\label{pba1}
P_{\bz}\left[X\right] = \pr{X} + \mathrm{Cov}\left(X, \bz\right)\var{\bz}^{\dagger}(\bz - \pr{\bz}),
\end{equation}
(where $\dagger$ denotes the Moore-Penrose generalised inverse). To form the required Hilbert space and perform the PBA requires direct elicitation of $\pr{X}$, $\pr{\bz}$ $\var{X}$, $\var{\bz}$ and $\var{X-\Ex{X \mid D, M_i}}$ for each $i$ (from which we can derive the quantities required to calculate \eqref{pba1}). 
Note that \eqref{pba1} represents a particular model weighting and so can be considered to be a model-synthesis method.

Given that the inner product and norm of our Hilbert space is a prevision, the distance between two objects in this space is equivalent to the modeller's uncertainty regarding their difference (or the extent to which they believe them to be the same). For example, when $X$ and $P_{\bz}(X)$ are scalar quantities, $||X-P_{\bz}(X)||^2=\Var(X-P_{\bz}(X))$. The distance between $X$ and $P_{\bz}(X)$ is therefore equivalent to the modeller's uncertainty as to the difference between their PBA and the `truth', which is minimised via \eqref{pba1}. 

Suppose we will observe our data and the results of our models by time $t$. Let $P_t(X)$ to denote primitive expectation at this time $t$. As argued by \cite{goldstein1997prior}, true belief at time $t$ need not coincide with the posterior of any particular Bayesian analysis (even if the model were believed) and the same is true for the PBA \citep{PBA}. Under relatively weak assumptions (see \cite{PBA},\cite{goldstein1997prior} for details), we can obtain
\begin{equation}\label{orthog1}
    (X - \Ex{X \mid D, M_i}) = (X-P_t(X))\oplus(P_t(X)-\Ex{X \mid D, M_i})
\end{equation}

\begin{equation}\label{orthog2}
    (X - P_{\bz}\left[X\right]) = (X-P_t(X))\oplus(P_t(X)-P_{\bz}\left[X\right])
\end{equation}

\noindent where the operator $\oplus$ indicates the sum of orthogonal components. \citep{PBA} utilise \eqref{orthog1} and \eqref{orthog2} to show that \eqref{pba1} is closer in mean-squared error to $P_t(X)$ than any individual $\Ex{X \mid D, M_i}$. \cite{PBA} proceed to partition the, in principle infinite, set of posterior expectations from models and priors that we might fit, into a finite collection of co-exchangeable classes $\bc=(C_1,...,C_m)$. The co-exchangeability implies that any pair of posterior judgements within the same class are a priori exchangeable, and any pair from different classes have the same correlation. With this structure, \cite{PBA} showed how using the available $\Ex{X \mid D, M_i}$ to learn about the class means and using the updated means rather than $\bz$ in \eqref{pba1} satisfied the same properties as the original PBA in the infinite model case. 

\section{Posterior Belief Assessments on AI systems}\label{sec::pba2}

The prior judgements that the modeller is required to specify for a PBA shed light on aspects of model development (including human insight and/or intervention) that are often obscured when detailing the use of AI systems in decision-making processes. In particular, the direct elicitation of the needed prior quantities require the modeller to go beyond considerations of accuracy (in the sense of empirical performance on testing data given training data) when considering the reliability of any individual AI system. $\var{X-\Ex{X \mid D, M_i}}$, in particular, defines the modeller's beliefs as to the variability in the difference between the true value of the QoI and the estimate given by the AI system (if it is a Bayesian model). This is a quantity which highlights judgements as to how well each model works given the situation at hand as well as potential weaknesses and blind spots. Subsequently, via PBA, the ways in which the interpretation of the models rely on external judgements made by modellers are identified and highlighted, making it possible to evaluate the credibility of such judgements and their implications for the models, and thereby decreasing the opacity of the models. 

Additionally, PBAs do not require that the modeller believes the raw results produced by the systems themselves. What the modeller needs to believe is their own second-order judgements updated upon observing these results. This helps to clarify who bears responsibility for different aspects of the model, and how accountability may be apportioned if outputs turn out to be problematic or unreliable. In particular, accountability can then be justifiably assigned to the modeller(s) who completed the assessment, and subsequently whoever is willing to accept and act on this uncertainty in downstream decisions. 

We highlight these strengths of the methodology through our application in Section \ref{sec::casestudy} and further discuss the ethical implications in Section \ref{sec::ethical.discussion}. The rest of this section expands the theory of PBA in order to both deliver a practical inferential methodology for general AI systems, going beyond Bayesian models, and to establish some of the properties we will require for our ethical discussion in Section \ref{sec::ethical.discussion}.

\subsection{Theory}\label{theory}

Specific to the development and use of AI systems is the indifference between different modelling judgements that can be induced by the complexity of modelling frameworks. It is easy to imagine such indifference in instances such as the fitting of a Bayesian Neural Network (BNN). In regards to prior specification for BNN's alone, there is a broad literature and hence a lot of scope for user choice (\cite{Ridgelet},\cite{howgood},\cite{Nalisnick},\cite{Allyouneed}). We consider the existence of a continuum of such models, partitioned into co-exchangeable classes as described in Section \ref{sec::pba1}, allowing a modeller to capture the uncertainty that arises from observing a finite number of AI systems in a sea of alternative configurations.

We extend PBA beyond the synthesis of Bayesian models in the original development, to the synthesis of estimates from general AI-systems, for example, some of our models might report no uncertainty at all in their estimate for a QoI (most Neural Networks have this property). A general version of the orthogonal decompositions shown in \eqref{orthog1} and \eqref{orthog2}, 
\begin{equation}\label{orthog}
    (X - W) = (X-P_t(X))\oplus(P_t(X)-W), 
\end{equation}

\noindent holds for any random quantity, $W$, which will certainly be known to us at time $t$, if we assume our previsions to be conglomerable, i.e. if $P(P_t(X)) = P(X),$ \citep{goldstein2013observables}. We will use \eqref{orthog} to prove an analogous, but generalised, version of the optimality results in \citep{PBA} that apply to any AI generated estimate that the user will a) have observed by time $t$ and b) believes to be informative in regards to the quantity of interest. 

Before proceeding to this, we highlight the fundamental setup of our problem. Consider $m$ classes and $n_i$ individual members of class $i$. Let $\bz = \{Z_{ij}$\;;\;  $i= 1, \ldots, m; \; j= 1, \ldots, n_i\}$, where $Z_{ij}$ is the estimate of the QoI from the $j$th model in class $i$. The judgement of within-class exchangeability implies, via the representation theorem \citep{BL} that $$Z_{ij} = \mu_i + \R_{ij},$$ where vectors $\mu_i$ and  $\R_{ij}$ are uncorrelated and where the latter has prior expectation $0$. The exchangeability therefore implies that individual models within a class are informative for the class mean $\mu_i$. Note that $\Cov\left(Z_{ij},Z_{kl}\right) = \Cov\left(\mu_i,\mu_k\right)$ for $i\neq k$ and $\Cov\left(Z_{ij},Z_{im}\right) = \Var\left(\mu_i\right)$ for $j\neq m.$

The assumption of co-exchangeability between the quantity of interest, $X$, and the members of the different classes is satisfied via 
\begin{equation}\label{co-ex.assumption}
X = \sum_{i=1}^{m}A_i\mu_i + \bU.    
\end{equation}
Throughout the paper we utilise block matrices and vectorizations to condense notation. For example, we write $X=\am+\bU$ where $\bA = (A_1,...,A_m)$ and $\bm=\text{vec}\big((\mu_1, \ldots, \mu_m)\big)$. $\bA$ is a known matrix and $\bU$ is a model discrepancy term which is uncorrelated with all mean and residual elements. Note that $\bU$ is a quantity that we cannot learn upon observing our data $\bz$, meaning that it presents an opportunity for the modeller to define a proportion of the uncertainty in $X$ that cannot be resolved via observing the AI systems under consideration. We explain the necessary judgements to derive $\bA$ and the uncertainty in $\bU$ later in this section. 

Our first result uses \eqref{co-ex.assumption} to derive a particularly useful form for the PBA for the QoI, $P_{\bz}(X)$.

\begin{lemma}\label{pba_update}
$P_{\bz}(X) = P_{P_{\bz}(\bm)}(X)=\bA\;P_{\bz}(\bm) + \pr{\bU}$
\end{lemma}

Lemma \ref{pba_update} states that updating our beliefs about the quantity of interest by the data is equivalent to updating by the adjusted class means, which involves adjusting the prior class means by the data. The proof is in \textcolor{red}{B.1} of the Supplementary Material.

\begin{corollary}\label{sample_means}
    Denote the sample means by $\overline{\bz} = \left[\;\overline{Z}_1,...,\overline{Z}_m\;\right]$, where $$\overline{Z}_i = \frac{1}{n_i}\sum_{j=1}^{n_i}Z_{ij}.$$ It is sufficient to utilise the sample means from each co-exchangeable class $\overline{\bz}$ to update $\bm$ (i.e. $\lfloor \bz \indep \bm \rfloor \;/\; \overline{\bz}$).
\end{corollary}

This result is proved in \textcolor{red}{B.2} of the Supplementary Material. Note that the extension of this sufficiency to $X$ is immediate from Lemma \ref{pba_update}.

Lemma \ref{pba_update} and Corollary \ref{sample_means} offer flexibility to the modeller, enabling the PBA to be constructed in a number of ways. We advocate the updating of the class means by the sample means followed by the updating of $X$ by the class means. Updating by the sample means gives the freedom to simulate large numbers of outputs for each class whilst ensuring that the updates are still computationally feasible. Maintaining the intermediary modelling step of updating the class means and \textit{then} updating the truth, though computationally unnecessary, provides useful structure to the assessment of belief. We denote the PBA as $P_{\mathcal{Z}}(X)$ from this point onward, where $\mathcal{Z}=P_{\overline{\bz}}(\bm)$. 

The matrix $\bA$ and uncertainty judgments for $\bU$ can be derived from the modeller's prior specification. Given a specification of $P(X)$, $\Var(X)$, $\Cov(X,\overline{\bz})$, $P(\bm)$ and $\Var(\bm)$ (we present a detailed example for obtaining these quantities in practice in Section \ref{sec::casestudy}), it follows from \eqref{co-ex.assumption} that
\begin{align}\label{VarU}
\pr{X}=\pr{\bA\bm+\bU}=\bA\pr{\bm}+\pr{\bU},\qquad \Var(X)=\bA\Var(\bm)\bA^{\intercal}+\Var(\bU),
\end{align}

\begin{equation}\label{A}
\Cov(X,\overline{\bz})=\Cov(X,\bm)=\bA\Cov(\bm,\bm)=\bA\Var(\bm).
\end{equation}

Rearranging (\ref{A}) gives $\Var(\bm)\bA^{\intercal}=\Cov(\bm,X)$. By \textcolor{red}{A.5} in the Supplementary Material, solving for $\bA^{\intercal}$ gives $$\bA^{\intercal}=\Var(\bm)^{\dagger}\Cov(\bm,X) + (\boldsymbol{I}-\Var(\bm)^{\dagger}\Var(\bm))\boldsymbol{T},$$ where $\boldsymbol{T}$ is an arbitrary matrix. If $\Var(\bm)$ is full rank, then $\Var(\bm)^{\dagger}\Var(\bm)=\boldsymbol{I}$ by \textcolor{red}{A.2} in the Supplementary Material and 
\begin{equation}\label{derived_A}
\bA=\Cov(X,\bm)\Var(\bm)^{\dagger}.
\end{equation}
Hence, the judgements that specify the prior uncertainty in the class means and their covariance with the quantities of interest, are sufficient to determine $\bA$. Then, given $\bA$, the prior specifications for $P(\bU)$ and $\Var(\bU)$ can be calculated via \eqref{VarU}. We discuss possible extensions that allow $\bA$, $P(\bU)$ and $\Var(\bU)$ to be learned in the discussion.

Before proceeding, note that we abandon the concept of the `initial Bayesian analysis', which is utilised by \cite{PBA} as a benchmark against which the optimality of a PBA can be judged. \citep{PBA} specify that the user has the freedom to deem any model output as having come from the `initial Bayesian analysis'. However, this indifference is misplaced in the sense that the output which is selected is in fact treated differently to the other model outputs in the framework. For example, this output is not placed into a co-exchangeable class. We propose that it should be, as it may have come from a group of methods which we wish to learn from \textit{and} about. A generalised version of the optimality result can be proven regardless. 

\begin{theorem}\label{main}

i) The Posterior Belief Assessment $P_{\mathcal{Z}}(X)$ is at least as close to `the truth', $X$, as any observed result from a single AI system (i.e. any single element of $\bz$).\\
ii) The Posterior Belief Assessment $P_{\mathcal{Z}}(X)$ achieves i) purely by virtue of being at least as close to `true belief', $P_t(X)$, as any observed result from a single AI system (thus explicitly establishing PBA as an $\mathcal{M}$-open form of inference).\\

\noindent \textit{Proof.}\\

\noindent i) \begin{align}
    ||X-P_{\bz}(X)||^2&\leq||X-Z_{ij}||^2\Leftrightarrow \label{i1}\\ 
    ||X-P_{\mathcal{Z}}(X)||^2&\leq||X-Z_{ij}||^2. \label{i2}
\end{align} 

\noindent \eqref{i1} holds by the definition of adjusted expectation (see Section \ref{sec::pba1}). Each respective element of $\bz$, $Z_{ij}$, is a linear combination in $\bz$ which can be recovered via $P_{\bz}(X)$ given specific prior specifications. The step from \eqref{i1} to \eqref{i2} holds via Lemma \ref{pba_update} in combination with Corollary \ref{sample_means} (establishing that $P_{\bz}(X)$ is equal to $P_{\mathcal{Z}}(X)$). 

\noindent ii) \begin{align}
    ||X-P_{\bz}(X)||^2&\leq||X-Z_{ij}||^2\Leftrightarrow \label{ii1}\\ 
    ||X-P_t(X)||^2+||P_t(X)-P_{\bz}(X)||^2&\leq||X-P_t(X)||^2+||P_t(X)-Z_{ij}||^2 \Leftrightarrow \label{ii2}\\
    ||P_t(X)-P_{\bz}(X)||^2&\leq||P_t(X)-Z_{ij}||^2 \Leftrightarrow \label{ii3}\\
    ||P_t(X)-P_{\mathcal{Z}}(X)||^2&\leq||P_t(X)-Z_{ij}||^2 \label{ii4}
\end{align} 

\noindent \eqref{ii1} holds via i). The decomposition of \eqref{ii1} in \eqref{ii2} holds via \eqref{orthog} and the properties of inner products (see proof of Theorem 6.11 in \cite{inner}). The step from \eqref{ii3} to \eqref{ii4} holds via Lemma \ref{pba_update} in combination with Corollary \ref{sample_means} (establishing that $P_{\bz}(X)$ is equal to $P_{\mathcal{Z}}(X)$). 

\qed

\end{theorem}
Recall that \cite{Jack2} describe the subjective, $\M$-open framing of GBI to be that the DGP represents a user's true beliefs and uncertainties, with the model acting as an imperfect representation of these beliefs. Here, the PBA (and associated uncertainties) are our `model' in the sense that they are the means by which we aim to achieve closeness to the true belief, $P_t(X)$. Note that $||X-P_{t}(X)||^2$ is a portion of the uncertainty that is present and cannot be resolved via any inferential method. 

The final result we present concerns the convergence of the PBA as the number of AI systems observed in each of the respective classes tends to infinity.

\begin{lemma}\label{conv}
    $ P_{\mathcal{Z}}(X) \rightarrow \am + P(\bU)$ as $n_i \rightarrow \infty$ $\forall i$.
\end{lemma}

Lemma \ref{conv}, proved in \textcolor{red}{B.3} of the Supplementary Material, shows that as the number of AI systems we have evaluated in each co-exchangeable class tends to infinity, the PBA for $X$ converges to $\am + P(\bU) $. As $\bA$ is defined a priori, this implies that the ultimate weighting of the models across classes is user-specified. Unless $\bA$ is set to select a single model, the PBA does not converge to a single model, thus avoiding one of the main criticisms of Bayesian Model Averaging \citep{yao2018using}. Note that the discrepancy remains at the prior mean, $P(\bU)$, pointing to the fact that there is a piece of uncertainty in $X$ that is not resolved by the models we consider as part of our framework.  

\section{Case Study}\label{sec::casestudy}

\subsection{Introduction}\label{CovidIntro}

We now introduce a case study regarding major decisions that were made in response to the evolving state of the Covid-19 pandemic in England. The pandemic represented a scenario in which levels of expertise in respective areas varied drastically across each stage of the decision-making process. 
From an overly simplified perspective, the statisticians will have had knowledge of the modelling, the medical experts would have had knowledge of the disease and the government officials will have had knowledge of the wider implications of different policies. There is an element of trust whenever information is passed between these groups, as well as a responsibility to process said information from one's respective position of expertise and (ideally) a sense of reciprocal accountability between the different groups of experts involved. A review of this situation and the nuances it contained can be found in \cite{Sab2021}.

Our case study involves a group of statisticians from the University of Exeter that were asked to develop models which could be used to track variants of concern (in England) during the pandemic and to report their results forward to the Scientific Pandemic Influenza group on Modelling, Operational sub-group (SPI-M-O), a subgroup of SAGE - the Scientific Advisory Group for Emergencies. During the pandemic, scientists and modellers would, by invitation, report modelling/results/findings to SPI-M-O. SPI-M-O would then report key insights from the UK modelling effort to SAGE who, in turn, would synthesize this information with information from other subgroups in order to inform the UK government. The approach of this specific team, at the time, was to develop different models in parallel and to pass the respective results from each up to SPI-M-O, who were working with many other teams from a number of institutions. 

In this case study, we revisit one particular week of modelling near the beginning of the Omicron wave (December 8th, 2021) and look at how the modellers could have combined their models using a PBA in order to report their own uncertainties to SPI-M-O, rather than passing on a collection of model results and leaving the synthesis work to them. The aim of this study is to demonstrate the care required to report uncertainty through a PBA and to discuss the process through an Ethical AI lens. It is not our aim to present any results relevant to the pandemic, nor to critique the work of these specific statisticians, SPI-M-O or any other groups that were providing unpaid expertise and modelling effort during a very challenging time for the UK. That said, were we to be faced with similar challenges now, we would use the techniques developed below to synthesise models in order to report our uncertainty to SPI-M-O. We would also advocate use of these techniques to bodies like SPI-M-O as a principled and ethical way of synthesising models from different groups. 

\subsection{The Models}\label{CovidModels}

During the Omicron wave, the group at Exeter developed 4 statistical models, which we introduce here. The goal of this section is not to go into extensive detail as to how each of these models worked, but instead to gain a general sense of the different modelling choices that were being made. This will help to motivate discussion in the following sections as to how these differences played into the overall uncertainty captured by the PBA. Details regarding the spatial discretisations of England discussed below (i.e. Middle layer Super Output Areas (MSOA), Lower layer Super Output Areas (LSOA), regional), as well as the boundaries associated with these geographies, can be found at \cite{Boundaries,Portal}.

At this time, the absence of the S-gene was a useful indicator as to whether or not a case was of the Omicron strain in contrast to the dominant (at the time) Delta strain, characterised by the presence of the S-gene marker \citep{challen2021early}. The number of S-negative infections in each location was used as a proxy by the team for the number of Omicron cases.

Let $Y_{jt}$ represent the number of Omicron cases reported on day $t$ at location $j$. Our simplest model, represented in \eqref{Model11}, is a hierarchical Bayesian model applied to the 9 regions of the UK. 
\begin{equation}\label{Model11}
\begin{split}
    Y_{jt} &\sim \text{NegBinomial}(\mu_{jt},\phi)\\
    \log(\mu_{jt}) &= \beta_0 + {\beta_1}t + \beta_{0j} + \beta_{1j}t \\
    \beta_{ij} &\sim \mathrm{N}(0, \Sigma_i) \qquad i= 0,1.
    \end{split}
\end{equation}
Priors on the $\beta$'s and $\phi$ were relatively uninformative and, in the case of the regression coefficients had mean $0$. Model 2 \eqref{Model21} is a spatio-temporal Poisson generalised additive model (GAM), also fit directly on the number of S- cases. We write this as
\begin{equation}\label{Model21}
\begin{split}
     Y_{jt} &\sim \mathrm{Pois}(\mu_{jt}) \\
    \log \mu_{jt} &= \lambda_0 + f(\text{MSOA}_j, t),
\end{split}
\end{equation}
where $f(\text{MSOA}_j, t)$ is smooth function that is separable in space and time. The spatial component is a Gauss-Markov Random Field (GMRF) whilst the time component is a cubic spline. Note that this is a higher resolution statistical model, breaking down the 9 regions of England into 6856 (nested) MSOAs.

Models 3 and 4, rather than modelling the S-negative cases directly, were developed (during the Delta wave) to partition the total number of Covid cases across multiple variants of concern and, in part, motivated by accounting for recorded cases where the S-gene marker was not tested for. Re-purposed for Omicron, these models separately treated total Covid cases spatio-temporally, with the log-Poisson rates modelled in 2 different ways. Model 3 used a separable structure with a GMRF over MSOAs and a 2nd order random walk in time, whilst Model 4 modelled the log-Poisson rates using a 2-layer Deep Gaussian Process \cite{salimbeni2017doubly, ming2023deep} over the 33755 LSOAs (nested within the MSOAs) in England. Models 3 and 4 then used a Bernouilli model (at the same spatio-temporal resolution as their counts models) for the S-gene status with the logit of the probability that a case was S-negative (here a proxy for an Omicron case) following the same structure as the latent layer in their respective Poisson models (a separable GMRF-random walk for Model 3 and a 2 layer Deep Gaussian Process for Model 4). In both cases, the expected number of Omicron cases can be derived by combining the classifier and the counts models. As the purpose of this case study is to examine the process of synthesising the model output via PBA, we don't provide the data and code for fitting the models, but we do provide the model output for each of the models, aggregated to the 9 regions used for our PBA, along with the code for fitting the PBA so that the results in this paper can be reproduced. 

\subsection{The PBA}

Our PBA focuses on the beliefs of the principal investigator of the group, who fit Model 1 and helped develop the approaches behind Models 3 and 4 (implemented by others in the group). They were responsible for communication between the group and researchers on SPI-M-O, both in the transmission of data and research questions from SPI-M-O to the group and of the group's results back to SPI-M-O. Note that due to the limited time and resources that were available while carrying out this kind of modelling on a weekly basis, the list of models included in this PBA is not an exhaustive list of the models that this group deemed to be viable given the problem at hand, but a recording of the models that were committed to under pressure. 

For the purposes of reporting results to SPI-M-O researchers, the main interest was in regional expected case numbers, so we focus this PBA at this resolution. The group also produced higher resolution maps to visually represent their insights to SPI-M-O, though we do not consider PBA for these. The resolution at which a PBA is completed should be reflective of the resolution at which a modeller believes themselves capable of specifying their beliefs. Our quantity of interest, $X$, will be the number of Omicron cases in each of the 9 UK regions on December 8th 2021. We note here that this PBA is taking place for illustrative purposes in 2023, with our modeller having to recall their thought processes for the date in question. In practice, this exercise should be carried out prior to results being communicated to decision makers.

\subsection{The Elicitation}\label{elicitation}

\subsubsection{$P(X)$ and $\Var(X)$}

Early December 2021 was a time of great uncertainty in regards to the Omicron variant. Uncertainty in $X$ must capture this uncertainty whilst capturing the modeller's judgements as to how the specific geography and infrastructure of England is affecting the spread of the virus. Two regions which are geographically distant may be well-connected through transport infrastructure whilst two regions that border each other may have populous cities that are, in fact, very far apart. Aspects such as this will, in turn, affect how likely it is that members of the public live in one region, but travel to another for work and/or leisure on a day-to-day basis.

We began by establishing whether there was an ordering of the regions that the modeller believed would best facilitate their thought process throughout this example. The modeller recalled that at this time, London was considered to be the epicentre of the Omicron outbreak and was thus believed to be `ahead of the curve'. In addition to this, the modeller felt that relative to the other regions, they were most confident in making their judgements in regards to how London was `connected' to the remaining eight regions. They therefore wished to specify their judgements regarding London first, so that they had the option to frame the remaining judgements around the situation there. The rest of the running order was determined by considering which region the modeller felt they would be ready to judge next, given the regions they had specified beliefs on up to that point. The order was: London, South East, North West, East, East Midlands, West Midlands, Yorkshire, North East and South West. 

$P(X)$ was the simplest specification to be made in this example. The modeller was confident in specifying values for this expectation directly: $P(X) = (400,180,150,80,80,40,40,40,20)^{\intercal}$. We note here that the time since the Omicron wave and the specificity of the date mean that this uncertainty is really a question of memory. In a real-time PBA, these numbers would be elicited prior to seeing the data for this week's cases but with knowledge of the previous week's analyses.

Eliciting coherent covariance/variance matrices directly is often not trivial (often Wishart or LKJ distributions are used). 
To ensure a coherent specification without the restrictions made by those distributional assumptions, we used an augmented version of the the method proposed by Kadane \citep{kadane1980interactive}. This method elicits a valid correlation matrix by eliciting a set of conditional previsions (based on hypothetical values of some elements of $X$) that can be used to derive the correlation matrix by solving a linear system of equations (full details are given in Appendix \textcolor{red}{C} in the Supplementary Material).

Beginning in London, the modeller had already specified $P(X_{\text{L}})=400$ and was then asked to specify $\Var(X_{\text{L}})$. The modeller gave $\Var(X_{\text{L}})=40,000$ arguing that, at this point in time, it was believed that cases were doubling in a matter of \textit{days} (\cite{PlanB} confirm that in early December 2021, it was believed that `cases could be doubling at a rate of as little as $2.5$ to $3$ days'.) The modeller deemed it within the scope of possibility that their $P(X)$ could wrong by `a couple of days' in all regions and therefore specified a doubling period as a two-standard deviation event which, in the case of London, gave a standard deviation of $200$. For the next region (South East), the Kadane method requires $P(X_{\text{SE}}|{X_{\text{L}}}=N_L)$ and $\Var(X_{\text{SE}}|{X_{\text{L}}}=N_L)$, where $N_L$ is a hypothetical number of cases that is chosen to be consistent with the prior specification for $X_L$. We chose $N_L = 500$ (\cite{kadane1980interactive} suggests using the mean + 1 standard deviation) and were given $P(X_{\text{SE}}|{X_{\text{L}}}=500)=200$ and $\Var(X_{\text{SE}}|{X_{\text{L}}}=500)=5625$. 

The method proceeds iteratively, e.g. now, for the North West, requiring $P(X_{\text{NW}}|{X_{\text{L}}}=N_L, X_{\text{SE}}=N_{\text{SE}})$ and $\Var(X_{\text{NW}}|{X_{\text{L}}}=N_L, X_{\text{SE}}=N_{\text{SE}})$. Once these conditional previsions and variances are collected for all regions (the final one being for $X_{\text{SW}}$ given consistent values for all other regions), we used the equations in \cite{kadane1980interactive} to derive the modeller's correlation matrix for $X$, which we then scale by the prior marginal variances (via \cite{al1998elicitation} and discussed in Appendix \textcolor{red}{C} of the Supplementary Material) to obtain the covariance matrix given in Table \ref{varX}.

\begin{table}[h!]
\begin{center}
\begin{tabular}{ c|c|c|c|c|c|c|c|c|c } 
         
            &
         \textbf{L} &
         \textbf{SE} & 
         \textbf{NW} & 
         \textbf{E} & 
         \textbf{EM} &
         \textbf{WM}  &
         \textbf{Y}  &
         \textbf{NE} &
         \textbf{SW} \\
\hline
\textbf{L} & 40000.00 & 8470.59 & 12026.76 & 1659.12 & 2614.73 & 809.40 & 685.99 & 685.99 & 357.60  \\ 
\hline 
\textbf{SE} & 8470.59 & 8100.00 & 4138.62 & 2327.65 & 1937.97 & 332.09 & 281.46 & 281.46 & 132.52  \\ 
\hline 
\textbf{NW} & 12026.76 & 4138.62 & 5625.00 & 1330.26 & 1310.28 & 608.41 & 515.64 & 446.89 & 136.19  \\ 
\hline 
\textbf{E} & 1659.12 & 2327.65 & 1330.26 & 1600.00 & 1084.54 & 209.83 & 177.84 & 163.61 & 35.60  \\ 
\hline 
\textbf{EM} & 2614.73 & 1937.97 & 1310.28 & 1084.54 & 1600.00 & 502.64 & 377.42 & 224.21 & 47.53  \\ 
\hline 
\textbf{WM} & 809.40 & 332.09 & 608.41 & 209.83 & 502.64 & 400.00 & 289.77 & 147.83 & 66.07  \\ 
\hline 
\textbf{Y} & 685.99 & 281.46 & 515.64 & 177.84 & 377.42 & 289.77 & 400.00 & 276.47 & 56.42  \\ 
\hline 
\textbf{NE} & 685.99 & 281.46 & 446.89 & 163.61 & 224.21 & 147.83 & 276.47 & 400.00 & 37.41  \\ 
\hline 
\textbf{SW} & 357.60 & 132.52 & 136.19 & 35.60 & 47.53 & 66.07 & 56.42 & 37.41 & 100.00  \\ 
\end{tabular}
\end{center}
\caption{The elicited prior variance matrix for the number of Omicron cases, $X$.}\label{varX}
\end{table}

\subsubsection{Model beliefs and their link to reality}

The modeller considered 3 co-exchangeable classes for the models. The first containing Model 1 and other (potential) hierarchical Bayesian regressions at higher resolution, as well as infinitely many `reasonable' perturbations to prior hyperparameters and the Hamiltonian Monte Carlo tuning parameters that might lead to slightly altered expectations following the analysis. Though only the data for Model 1 is available from this class, the modeller did, in later weeks, fit MSOA versions of these models (so in principle we could have sampled other variants from Class 1). Class 2 contained Models 2 and 3 as well as, in principle, perturbations to the prior hyper-parameters for their GMRF and separate temporal structures. Class 3 contains Model 4 as well as any variants of Deep Gaussian Processes with different numbers of inducing points, different start points for the optimisation in the variational inference, perturbations to the base correlation lengths for the Gaussian Process layers and so on. Note that these judgements do not amount to the models themselves being co-exchangeable, but rather the expectations for the QoI determined by the models being co-exchangeable (a priori). Following \eqref{co-ex.assumption}, $X = {A_1}\mu_1 + {A_2}\mu_2 + {A_3}\mu_3 + \bU,$ where the mean for class $i$ is $\mu_i$, $\bU$ is the discrepancy term and the $A$ matrices will be derived from the prior specification below as described in Section \ref{theory}.

The modeller judged the models in the 3 classes to be a priori unbiased, setting $P(\mu_1) = P(\mu_2) = P(\mu_3) = P(X)$. The final terms required were $\Var(\bm)$, describing how models within each class might vary and are related to each other, and $\Cov(X,\bm)$, describing how the output from the models was informative for the true number of Omicron cases. These were elicited using models from Class 1 as a baseline as these were the specific models developed and run by our modeller and are also the simplest and most interpretable. In a general setting, a baseline is not necessary, but it can be helpful for the modeller to frame their judgements.

We should expect all models to be positively correlated with $X$ (or they would not be deemed useful in the first place) and we should expect a reasonable level of correlation given that each model has been updated with the data and we are considering estimators for a QoI (in our case posterior expectations) and their correlation with the QoI itself. As Class 1 models have no explicit spatial correlation structure the modeller began by assigning a correlation of $0.75$ between the Class 1 mean and reality for each region, as a baseline (Table \ref{table:elicitation1}, row 4). Correlations between reality and other classes were specified for each region by the modeller upon considering whether the models in that class had particular strengths/weaknesses in relation to the baseline models.

The specified correlations between Class 2 models (with GMRF spatial components) are given on row 5 of Table \ref{table:elicitation1}, and reflect a number of qualitative judgements given by the modeller. The models are fit at MSOA level and are most reliable in regions with a large number of MSOAs that are close to each other and where the infection is more dominant within the region (rather than on the border of other regions) due to the aggregation of the reported results to region. In the modellers view, this made the approaches more accurate in London and NW and less accurate in regions with a lot of coastline like the SW, NE and WM (though landlocked, Wales is not in the data and so acts as a neighbourless border for the GMRF). The modeller also judged that regions neighbouring Omicron `hotspots' might be over-estimated due to the neighbours, hence they reduced their correlations between these models and reality in SE. 

The DGPs in Class 3 are designed to be able to overcome the rigidity of the neighbourhood correlations seen in the GMRF through the deep layers. They were thus given higher correlations with reality than most classes, other than WM, NE and SW where the large data-free borders provided by Wales and the coast were also judged to impact accuracy more than for Class 1 (see Table \ref{table:elicitation1}, row 6). The modeller also felt that data sparsity in SW, WM and NE might make the DGPs harder to identify and thus be potentially worse than the models in Class 2, particularly given that the models were fit using the smaller LSOAs rather than MSOAs. 

\begin{table} 
\begin{center}
\begin{tabular}{ c|c|c|c|c|c|c|c|c|c } 
& \textbf{L} & \textbf{SE} & \textbf{NW} & \textbf{E} & \textbf{EM} & \textbf{WM} & \textbf{Y} & \textbf{NE} & \textbf{SW}\\
\hline
 $\text{Corr}(\mu_1,\mu_2)$ & 0.70 & 0.75 & 0.75 & 0.75 & 0.75 & 0.75 & 0.75 & 0.75 & 0.65 \\
\hline
$\text{Corr}(\mu_1,\mu_3)$ & 0.70 & 0.75 & 0.75 & 0.75 & 0.75 & 0.75 & 0.75 & 0.75 & 0.65 \\
\hline
$\text{Corr}(\mu_2,\mu_3)$ & 0.75 & 0.75 & 0.75 & 0.75 & 0.75 & 0.75 & 0.75 & 0.75 & 0.75 \\
\hline
$\text{Corr}(X,\mu_1)$ & 0.75 & 0.75 & 0.75 & 0.75 & 0.75 & 0.75 & 0.75 & 0.75 & 0.75\\
\hline
$\text{Corr}(X,\mu_2)$ & 0.85 & 0.70 & 0.80 & 0.75 & 0.75 & 0.70 & 0.75 & 0.70 & 0.60\\
\hline
$\text{Corr}(X,\mu_3)$ & 0.85 & 0.80 & 0.80 & 0.75 & 0.75 & 0.65 & 0.75 & 0.65 & 0.60\\
\end{tabular}
\end{center}
\caption{Rows 1-3: Diagonal elements of the correlation matrices between distinct elements of $\bm$. Rows 4-6: Diagonal elements of the correlation matrices between $X$ and the elements of $\bm$.}\label{table:elicitation1}
\end{table}

The modeller was then asked to give regional correlations between the means for each class. The values given are shown in Table \ref{table:elicitation1} rows 1-3. The class means were considered to be relatively highly correlated across regions, with a value of $0.75$ assigned to the majority of the pairings. The modeller felt that the strengths of the Class 2 and 3 models in London and the perceived weaknesses of both in the SW would lead those classes to be more similar in those regions. 

We now obtain the standard deviation specifications needed to convert these correlations into variances/covariances. For the purposes of this example, we derived the forms of $\Var(\mu_i)$ and $\Var(\R_{ij})$ (for $i=1,2,3$) via utilising the same correlation structure as $\Var(X)$, but scaling the associated standard deviations. The modeller specified that they believed that Class 1 would be associated with the lowest variability while Class 3 would be associated with the highest (due to it being fit at the high-resolution LSOA level). To reflect this ranking, the modeller assigned $\Var(Z_1)$ $70\%$ of the variation in $X$, $\Var(Z_2)$ $85\%$ and $\Var(Z_3)$ $95\%$.Assigning the data in each class lower variability than $X$ is a reasonable assumption given that we are setting this example at a time in the pandemic of particularly high uncertainty, coupled with the fact that our modeller was a statistician as opposed to a medical professional.

Within each class, this uncertainty was then partitioned (in terms of $\Var(X)$) to determine what proportion would be attributed to the respective mean and residual components. $\Var(\mu_i)$ (for i=1,2,3) was set to be a relatively high proportion of $\Var(X)$, in order to reflect the fact that there were very few samples from each class. The remaining uncertainty was then assigned to $\Var(\R_{ij})$. This quantity was `highest' for models in Class 3, due to the variability as to where the inducing points were placed between individual model runs, and lowest in Class 1, due to the fact that the modeller believed the majority of the individual variability here would be down to relatively small variation in the MCMC sampling that took place during fitting (see Table \ref{table:elicitation2}). Note that $\Var(\overline{Z}_1)=\Var(Z_1)$ and $\Var(\overline{Z}_3)=\Var(Z_3)$, as we have a single observation from Class 1 and 3 respectively. As a result of the fact that we have two observations from Class 2, $\overline{Z}_2 = \mu_2 + \frac{1}{2}\sum_{j=1}^{2}\R_{2j}$ and it follows that $\Var(\overline{Z}_2)=\Var(\mu_2) + \frac{1}{2}\Var(\R_{2j})$. 

\begin{table}
\begin{center}
\begin{tabular}{ c|c|c|c|c|c|c } 
 & $\Var(\mu_1)$ & $\Var(\R_{1j})$ & $\Var(\mu_2)$ & $\Var(\R_{2j})$ & $\Var(\mu_3)$ & $\Var(\R_{3j})$ \\
\hline
\text{$\%$ of $\Var(X)$} & 65 & 5 & 75 & 10 & 80 & 15
\end{tabular}
\end{center}
\caption{The scaling of $\Var(X)$ used to obtain $\Var(\mu_1)$, $\Var(\mu_2)$, $\Var(\mu_3)$, $\Var(\R_{1j})$, $\Var(\R_{2j})$ and $\Var(\R_{3j})$. }\label{table:elicitation2}
\end{table}

The off-diagonal elements of $\Cov(\mu_i,\mu_j)$ (where $i\neq j)$ and $\text{Cov}(X,\mu_i)$ (for $i=1,2,3)$ are not as easy to consider. It is not immediately obvious, for example, as to how $\mu_{1\lo}$ should relate to $\mu_{2\se}$. These are complex judgements that the modeller may not feel able to confidently specify, in which case one can employ the simplification of setting these off-diagonal entries to zero, thus solely relying on the judgements that have been made between corresponding regions (such as those in Table \ref{table:elicitation1}). In an example in which all covariances are positive, as in our application, setting the off-diagonal entries to zero would lead to a conservative Posterior Belief Assessment, as it would restrict the extent to which learning could be done between regions and/or classes. This would be arguably preferable to producing an overly confident PBA, especially when considering such ethically complex scenarios. 

For this application, we demonstrate the use of belief separations in order to derive these specifications from judgements that have already been made. A belief separation is an orthogonality judgement between two quantities given a third. For example, we might invoke a belief separation that $\mu_{il}$ and $\mu_{jm}$ are orthogonal given $\mu_{jl}$ (we use the convention that $i$, $j$ here index the class and $l$, $m$ the regions). In words, and using a particular example, this might mean $\mu_{2\se}$ is considered to be useful in learning about $\mu_{1\lo}$ only in terms of what it says about $\mu_{2\lo}$ (which in turn will teach you about $\mu_{1\lo}$). The above belief separation implies $\Cov(\mu_{il},\mu_{jm})=\Cov(\mu_{il},\mu_{jl})\Var(\mu_{jl})^{\dagger}\Cov(\mu_{jl},\mu_{jm})$  (see Section 5.15 of \cite{BL}). Note that we have already elicited all terms on the right hand side of this relationship, so, given appropriate belief separation judgements, we can simply calculate the required cross-covariances between different regional means in different classes (via \textcolor{red}{A.1} in the Supplementary Material). To ensure valid covariance matrices, we make all necessary belief separations for $j>i$ and $m>l$, thus populating the upper triangle of the covariance matrix, which can then be reflected to obtain all covariances. Similarly, for the off-diagonal elements of $\text{Cov}(X,\mu_i)$ (for $i=1,2,3)$, we repeated this process utilising the separation $\Cov(X_{l},\mu_{jm})=\Cov(X_{l},\mu_{jl})\Var(\mu_{jl})^{\dagger}\Cov(\mu_{jl},\mu_{jm})$. 

Note that, similarly to most of the quantities considered in this framework, belief separations are a subjective prior judgement which can be defined in a number of ways. When building $\Cov(\mu_i,\mu_j)$ (where $i\neq j)$ we made a specific simplifying assumption. $\mu_{il}$ and $\mu_{jm}$ would, in theory, intuitively be separated by both $\mu_{jl}$ \textit{and} $\mu_{im}$. However, we assume that each pair is separated by one specific element only (i.e. $\mu_{jl}$ \textit{or} $\mu_{im}$). The single elements that were used were determined by the modeller's preference ordering over regions (i.e. the London specific elements were used most frequently whilst the South West specific elements were used least frequently). Proceeding in this way, with this level of complexity, was sufficient for the purposes of this example. 

\subsection{Results}\label{sec::results}

Given the elicited quantities in Section \ref{elicitation}, we can calculate $\bA$, $P(\bU)$ and $\Var(\bU)$ (via \eqref{VarU} and \eqref{A} respectively) and then derive our Posterior Belief Assessment via \eqref{pba1}\footnotemark[3]. We obtained $$P_{\mathcal{Z}}(X)=(1125.01,483.28,370.83,333.38,144.87,50.83,65.88,8.97,60.42)^{\intercal}.$$ 

\footnotetext[3]{$\bA$ and $P(\bU)$ are included in Appendix \textcolor{red}{D} of the Supplementary Material. The diagonal elements of $\Var(\bU)$ are included in Table \ref{table:elicitation4}.}

\begin{table}
\begin{center}
\begin{tabular}{ c|c|c|c|c|c|c|c|c|c } 
& \textbf{L} & \textbf{SE} & \textbf{NW} & \textbf{E} & \textbf{EM} & \textbf{WM} & \textbf{Y} & \textbf{NE} & \textbf{SW}\\
\hline
$P_{\overline{\bz}}(\mu_1)$ & 965.63 & 312.17 & 277.03 & 300.12 & 113.57 & 61.13 & 68.93 & 7.79 & 51.39\\
\hline 
$P_{\overline{\bz}}(\mu_2)$ & 1012.51 & 382.49 & 292.58 & 336.13 & 143.77 & 81.31 & 83.50 & 18.19 & 63.19\\
\hline 
$P_{\overline{\bz}}(\mu_3)$ & 1067.25 & 401.51 & 300.15 & 337.52 & 144.90 & 83.56 & 86.26 & 22.80 & 74.63\\
\hline
$P_{\mathcal{Z}}(X)$ & \textbf{1125.01} & \textbf{483.28} & \textbf{370.83} & \textbf{333.38} & \textbf{144.87} & \textbf{50.83} & \textbf{65.88} & \textbf{8.97} & \textbf{60.42}\\
\end{tabular}
\end{center}
\caption{Rows 1-3: The adjusted expectations of the elements of $\bm$ by $\overline{\boldsymbol{Z}}$. Row 4: The Posterior Belief Assessment $P_{\mathcal{Z}}(X)$.}\label{table:elicitation3}
\end{table}

\begin{figure}[hbt!]
     \centering
     \begin{subfigure}[b]{1\textwidth}
         \centering
         \includegraphics[width=\textwidth]{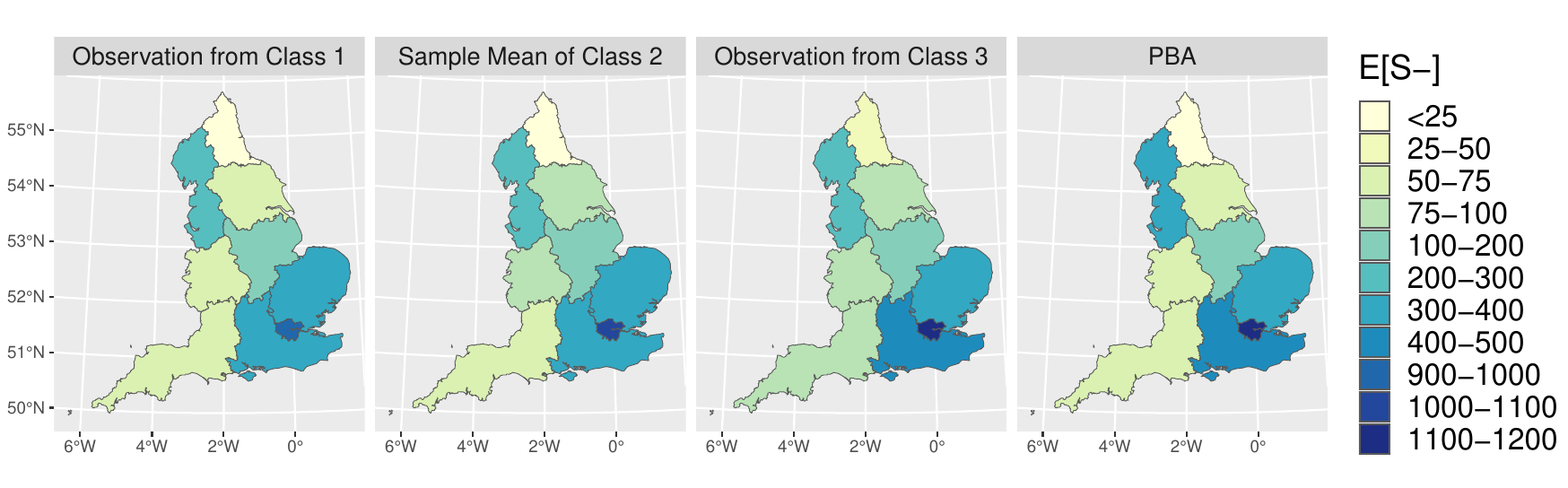}
     \end{subfigure}
        \caption{The PBA and elements of $\overline{\bz}$ plotted on a regional map of England. The digital boundaries used to produce this plot were obtained from \citep{Portal}  (Source: Office for National Statistics licensed under the Open Government License v.3.0. Contains OS data \textcopyright\; Crown copyright and database right 2023). }
        \label{maps}
\end{figure}

Figure \ref{maps} compares the PBA with the model observations. Overall, the PBA seems to reflect the observation from Class 1 most closely, excluding London and its surrounding regions. It mirrors the level of cases in London from the observations in Classes 2 and 3, and the behaviour in the surrounding regions from the observation in Class 3. As expected, the observation from Class 3 exhibited the greatest differences between London and the East and the South East respectively. In addition to this, it modelled the greatest difference between the East and South East, indicating that the levels in these regions were allowed to decrease by different amounts. Similarly, in the PBA, the levels in London drop off by $791.63$ into the East and $641.73$ into the South East. Note that the expected case levels in the North West are highest in the PBA, this highlighting the modeller's belief that the North West and London are more connected than any of the models have captured.

\begin{figure}[hbt!]
     \centering
     \begin{subfigure}[b]{1\textwidth}
         \centering
         \includegraphics[width=\textwidth]{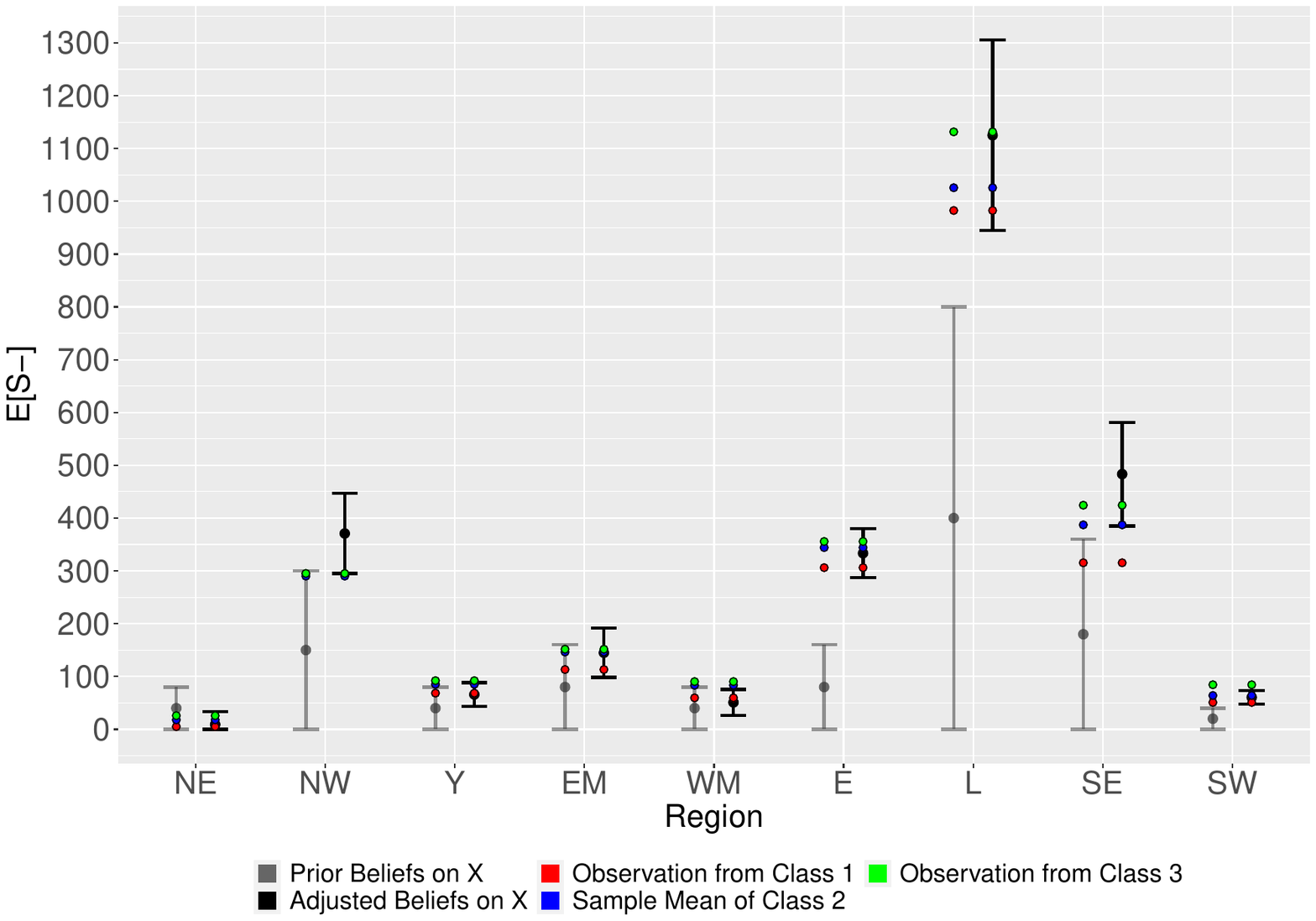}
     \end{subfigure}
        \caption{The elements of $\overline{\boldsymbol{Z}}$ plotted against $P(X)\pm2\sqrt{\Var(X)}$ and $P_{\mathcal{Z}}(X)\pm2\sqrt{\Var_{\mathcal{Z}}(X)}$ by region. Note that the lower boundary for the adjusted beliefs on NE was projected up to $0$, as in this case the $\pm2$ standard deviation range contained negative values.}
        \label{errors}
\end{figure}

The prior specifications in Figure \ref{errors} show that, on average, our modeller was initially underestimating the magnitude of the situation relative to the model observations. However, the adjusted beliefs on $X$ which consist of the PBA and the adjusted variance, 
\begin{equation}\label{adjvar}
    \Var_{\mathcal{Z}}(X) = \Var(X) - \Cov(X,\mathcal{Z})\Var(\mathcal{Z})^{\dagger}\Cov(\mathcal{Z},X),
\end{equation}
show that this was rectified upon updating by the data\footnotemark[4]. 
\footnotetext[4]{See \cite{PBA} for a similar example in which the adjusted variance is calculated and interpreted in light of the completed PBA.}

The adjusted beliefs plotted in Figure \ref{errors} are of particular interest as they not only show how the modeller's prior judgements have changed, but what elements of their judgements remain in light of the data. It can be seen more clearly (relative to Figure \ref{maps}) that the modeller believes that the model observations are underestimating the situation in the North West. As well as this, they believe that Model 1 (i.e. the observation from Class 1) predicted a value that was significantly too low for the South East. The only model that falls within the uncertainty bounds of the West Midlands is Model 1, reflecting the belief that performance in this region was relatively poor on the whole due to the fact that this region borders Wales, which was not included in the models. 
\begin{table}
\begin{center}
\begin{tabular}{ c|c|c|c|c|c|c|c|c|c } 
& \textbf{L} & \textbf{SE} & \textbf{NW} & \textbf{E} & \textbf{EM} & \textbf{WM} & \textbf{Y} & \textbf{NE} & \textbf{SW}\\
\hline
$\Var_{\mathcal{Z}}(X)$ & 8126.92 & 2408.02 & 1439.62 & 534.23 & 540.84 & 148.93 & 123.47 & 150.42 & 42.58\\
\hline 
$\Var(X)$ & 40,000 & 8100 & 5625 & 1600 & 1600 & 400 & 400 & 400 & 100\\
\hline
$\Var(\bU)$ & 6525.97 & 1822.21 & 1196.88 & 477.39 & 490.28 & 122.49 & 99.57 & 132.68 & 39.29\\
\hline
\text{RU} & $79.7\%$ & $70.3\%$ & $74.4\%$ & $66.6\%$ & $66.2\%$ & $62.8\%$ & $69.1\%$ & $62.4\%$ & $57.4\%$\\
\hline
\text{MRU} & $83.7\%$ & $77.5\%$ & $78.7\%$ & $70.2\%$ & $69.4\%$ & $69.4\%$ & $75.1\%$ & $66.8\%$ & $60.7\%$ 
\end{tabular}
\end{center}
\caption{Row 1: The adjusted variance on $X$, $\Var_{\mathcal{Z}}(X)$. Row 2: Diagonal elements of $\Var(X)$, given again here for computational ease. Row 3: Diagonal elements of $\Var(\bU)$. Row 4: The percentage resolved uncertainty in each region. Row 5: The max percentage of uncertainty in each region which could be resolved via observation of the outputs of Models 1-4.}\label{table:elicitation4}
\end{table}

Table \ref{table:elicitation4} shows how much of our prior uncertainty in $X$ we have resolved via the completion of the PBA. 
The elicited beliefs given imply prior beliefs on the quantity $\bU$ which means that the modeller has defined the proportion of $\Var(X)$ that it is \textit{possible} to resolve by observing models from the given classes. This upper bound can be found by taking $\left(1-(\Var(\bU)/\Var(X))\right)\times 100$. Then, the percentage of variation we have actually resolved in $X$ can be found via $\left(1-(\Var_{\mathcal{Z}}(X)/\Var(X))\right)\times 100$. For example, let us consider the fact that we have resolved $57.4\%$ of the variation in South West. According to the prior specifications of the modeller, the amount of uncertainty that \textit{can} be resolved in this direction is $60.7\%$, which coheres with the modeller's beliefs that these models would struggle to be sufficiently informative in this area.

Note that although our QoI was multidimensional, it still represented the value that had been explicitly modelled by the modelling team (i.e. $X$ was the number of cases and each of the models was a model for $X$ updated by covid case data). Our QoI need not have been the number of cases, but could have been a probability of cases exceeding some threshold or a policy relevant $R_t$ number. The models themselves would remain the same (modelling covid cases) and the model-QoI would be determined from each model with the PBA only considering judgements about the QoI for the models and reality.

\section{Ethical Implications}\label{sec::ethical.discussion}

When discussing ethical AI in this paper, we specifically concern ourselves with decisions made by statisticians and data scientists in the course of their everyday modelling practice, which shape what the results of AI systems truly mean and the ways in which they should be communicated to users. Such decisions are often perceived as being purely scientific and not part of ethical reasoning \citep{Sab2021}, whereas we argue that they play both roles at once: first, because while typically grounded in scientific reasoning, a technical assumption or interpretation is often likely to have wider-ranging repercussions, even if these are not immediately apparent to the modeller; and second, because within a decision-making process, technical judgements are intertwined with a wider cast of judgements and interpretation by a variety of experts, each of which will have their own respective skills, goals, background knowledge and related ethical concerns (see also \cite{leonelli2021data}, 84ff and 110ff; \cite{bezuidenhout2021does}). 

Our approach focuses on methods and strategies that statisticians and data scientists can adopt, within their own practice, to help tackle ethical concerns, particularly those linked to the opacity of model outputs, their use, and accountability for modelling practices that inform decision making. We argue that these concerns follow naturally from considerations regarding the statistical foundations that underpin the development and use of an AI system. Attention to the meaning and use of uncertainty within an AI system may not address all the ethical concerns around the potential misuse of such models, but it should be an important component in any debate around the ethical use of a particular system, and one that only modellers are in a position to offer. Extending the repertoire of strategies available to modellers to implement ethical reasoning is crucial, in our view, towards rooting ethical AI in actionable frameworks with tangible goals. This distinction was also adopted by \cite{saltz2019data}, who excluded papers that focused on `high-level societal ethical considerations' with the purpose of focusing on sources that discussed `actionable ethics analysis' by data scientists. 

The idea that accountability cannot be attributed to the modeller due to the complexities of AI systems hinges on the assumption that one cannot be responsible for a system that they do not understand, or a system that may result in unanticipated outcomes \footnotemark[5] \footnotemark[6]. However, we argue that accountability follows naturally from the modeller being a key contributor to the  development of AI systems, and thereby bearing some responsibility as to their functioning. Responsible practice does not mean that a  modeller has the ability to fully anticipate the consequences of deploying a system, which is never feasible for any technical innovation; rather, it means being reflexive and as open as possible about the influence that modellers did have on the development of such systems and the ways in which AI systems remain uncertain upon subsequent uses \footnotemark[7]. By requiring judgements a priori (i.e. before the outputs of the AI models are observed and a decision is made), PBAs prevent the modeller from altering their judgements depending on whether or not they wish to be perceived as being in (dis)agreement with the models (an issue which is referred to in the elicitation literature as \textit{hindsight bias} \citep{kadane1998experiences}). 

\footnotetext[5]{See \cite{novelli2023accountability} for a recent take on the definition of accountability in AI and a review as to why it has been historically difficult to assign, in part due to the opacity of the systems and the interdisciplinary nature of their use.}
\footnotetext[6]{\cite{martin2019ethical} acknowledge that assigning responsibility and accountability to those who develop algorithms (a general term they assume encapsulates many methods including `machine learning, artificial intelligence and neural networks') is often deemed to be `inefficient and even impossible'.}
\footnotetext[7]{See \cite{martin2019ethical} for a similar take on this issue within the context of the use of algorithms in business - they go as far as to take the stance that developers who produce `inscrutable' systems should be held to a relatively \textit{greater} level of accountability than those who do not.}

The degree to which a modeller has had causal influence on particular outcomes and how this relates to accountability for decisions made using AI-systems they developed will vary on a case by case basis, depending on the institutional context and the type of AI system in question. Moreover, accountability from the legal perspective is often divorced from accountability at the technical level. For example, a modeller may not (strictly speaking) be liable for damages if the use of an algorithm commercialised by their company goes wrong, but they may be asked to intervene and help to rectify the situation via their technical expertise. The first instance considers accountability in a legal sense, whilst the latter treats it more as an ethical principle (while not mutually exclusive, see \cite{durante2022legal} for explanation as to the differences between ethical and legal principles and the importance of acknowledging such differences). Thus, in an ethically motivated sense, modellers cannot escape some degree of accountability, purely by virtue of being the entity who will understand the system and its limitations to the best degree, and have the capacity to implement technical solutions where useful. And the act of making the judgements required to complete a PBA demand that the modellers actively preempt and prepare for the instance in which they may be called upon for such insight in future.  

It is an important distinction that a PBA has the potential to improve decision making in this setting as it facilitates subjective uncertainty quantification, thereby reducing opacity, acknowledging responsibility and aiding accountability by illuminating key judgements made by experts and opening them up to scrutiny. It is not automatically assumed that imbuing the results of AI systems with expert judgement `improves' the results in respects outside of our scope of interest. In our example, our expert’s particular area of expertise was in discussing the strengths and limitations of the modelling. An alternative domain-expert, such as a medical doctor, would no doubt have had more well-informed specifications of $P(X)$ and $\Var(X)$. That being said, it would still be of great value for them to be able to see how these quantities were defined by the people advising them, compared to their own perception of the situation. 

The development of alternative methods which claim to share similar motivations to PBAs (such as those discussed in Section \ref{sec::m-open}) are indeed principled, and target areas of model development that should be of interest. However, we must be open to recognising what they are actually optimising. Optimisations that take the form of divergences between full probability distributions, which are parameterised by systems such a neural networks, are not designed to target belief ownership (even if one is claiming them to be minimising the distance to `true beliefs'). Similarly, model synthesis techniques which can be used over groups of AI models do not automatically narrow uncertainty on the `truth' just by virtue of considering multiple systems. It therefore follows that we should utilise objectives that are \textit{built} with ethics in mind, instead of trying to shoehorn responses to such concerns into methodology where it will never be feasible. 

We hope this discussion serves as an example of the ways in which our take on ethical AI, characterised primarily by statistical philosophy and technical theory, can fit into the wider discussion whilst also raising new and complex considerations that should be studied alongside more `mainstream' concepts. 
 
\section{Conclusions}\label{Future}

In this paper, we presented the argument that consideration of the statistical foundations that underpin the development of an AI system and the way in which these choices affect the meaning of the results and their subsequent use is an ethical imperative. We maintain that this is an issue that can only be addressed effectively through the technical expertise of data scientists and statisticians in conjunction with the interpretation and framing that can only be provided by ethicists and context specific domain-experts. From a modelling perspective, we believe that a natural route towards addressing our concerns is the use of subjectivist approaches, the traditional versions of which have been rendered practically impossible by the complexity of AI systems. We present advanced theory for Posterior Belief Assessment against the backdrop of recent work on Generalised Bayesian Statistics and model synthesis, which has been developed to fit elements of subjectivism within the landscape of modern technology. While this alternative theory has its advantages, we highlight the ways in which the theory of PBA is more so motivated by justifiable ethical practice within the context of decision making. We illustrate the use of our methodology on a case study regarding models used to advise the UK government on the spread of the Omicron variant of Covid-19 during December 2021. 

In our illustrative PBA for Omicron cases, we used a mixture of elicitation techniques. How easy the required quantities are to think about for modellers and how experts should be guided in thinking about them remains a rich topic that demands further attention. In particular, the specification of large variance matrices requires a modeller to consider complex relationships across classes, whilst ensuring that they have provided a coherent specification. We used the method of \cite{kadane1980interactive} for this purpose and leveraged belief separations, though other approaches have been proposed in the literature (for example, see \cite{farrow2003practical}). 

Given the necessary belief specification, we could use the PBA framework to ascertain how many different exchangeable AI-systems \textit{should} be evaluated within each class in order to resolve a predetermined portion of the expert's uncertainty. The ability to determine the value of additional evaluations and the extent to which the remaining uncertainty is resolvable would be useful for determining whether or not to use the inferences in decision making.

We showed that the precise linear combination of model class means representing a QoI and the properties of the model discrepancy are specified (albeit indirectly in our application) by a user and cannot be learned through running more AI systems from the co-exchangeable classes. However, in applications where data and inference proceeds sequentially (as was the case during Covid-19 passing weekly updated modelling results to SPI-M-O), the difference between a predicted QoI (say at time $t-1$) and the realised QoI at time $t$ could provide data (under suitable exchangeability judgements across time) capable of learning $\bA$ and $\bU$.  This remains an avenue for future research.

In this paper, we have primarily discussed $\mathcal{M}$-open inference methods as a route to meaning from a purely philosophical perspective and have championed the properties of PBA. Further research is required to compare to alternative $\mathcal{M}$-open techniques, and to reflect on the meaning and ethical considerations from the results of each approach.

Though we have provided code for calculating the required quantities in a PBA, the burden of expert elicitation, or simply of careful introspection of the modeller might make our approach seem unattractive. In particular, when so many AI technologies give answers with minimal human input or interaction, methodology that explicitly requires that human input to derive meaning, might be criticised as not `objective'. We view this as a strength rather than a weakness. If the work to establish ownership, for the modeller and user, of the key inferential or decision critical statements made by an AI system is deemed to be too difficult, we might ask whether one should ever use such a system and under what conditions is it safe. Our thesis is that, in order to use AI ethically, you (or a modeller you trust) are required to specify your beliefs as to what it will say, how it relates to alternative systems and the extent to which it is informative about reality. These judgements are fundamental to principled decision making. Though there are many other facets to the ethical use of AI, this one in particular, which requires foundational understanding and input from the modellers themselves has not yet been addressed within the literature to date to our knowledge. 

Finally, we again acknowledge that our own framing of this particular facet of ethical AI, is derived from our own foundational philosophical worldview, holding that uncertainty is not a measurable property of the world but of individuals. The meaning of uncertainty, of probability itself and hence of the estimates given by our AI-systems, is fundamental in our approach to the ethical interpretation and use of AI. Though we acknowledge other worldviews and other foundations without criticism, we maintain that the work of establishing meaning from the fundamentals (uncertainty and probability) through to AI-derived estimates for QoI's to support decision making, must be done and the meaning must be commonly understood (and explainable) as a pre-requisite to alternative ethical approaches under these different worldviews. 

\textbf{Acknowledgements.} Cassandra Bird was funded by EPSRC studentship EP/V520317/1. Danny Williamson (DW) was funded by EPSRC grants EP/V051555/1 and EP/Y005597/1. Sabina Leonelli was funded by the European Research Council (ERC) under the European Union's Horizon 2020 research and innovation programme (grant agreement No. 101001145). The authors would like to thank Xiaoyu Xiong, Ben Youngman and Stefan Siegert, who together with DW formed the statistical modelling team who provided the data and insight for our application in Section 4. We also thank Lachlan Astfalk for his insight into the geometric interpretation of the PBA update.

\section*{Appendix A}
This section contains results from matrix theory and Bayes linear statistics that are used to prove our Lemmas and theorems in the main text.

\textbf{A.1} \citep{BL} \textit{A,B,C are three belief structures.} $$\Cov(A,B)=\Cov(A,C)\Var(C)^{\dagger}\Cov(C,B)$$ \textit{is equivalent to the condition that} $\lfloor A\indep B\rfloor/C$. \\

\textbf{A.2} \citep{GIB} \textit{Let A be an $m\times n$ matrix (with real/complex entries). Then a) $A^{\dagger}A=I_n$ if and only if rank(A)=n and b) $AA^{\dagger}=I_m$ if and only if rank(A)=m.} \\

\textbf{A.3} \citep{BL} \textit{Second-order belief specifications over two collections B,D are finite and coherent if and only if the joint variance-covariance matrix}

    $$\Var \left( 
    \begin{bmatrix}
    B\\
    D
    \end{bmatrix}
    \right) =
    \begin{bmatrix}
    \Sigma_{B} & \Sigma_{BD}\\
    {\Sigma^{T}}_{BD} & \Sigma_{D}
    \end{bmatrix}$$

\textit{is non-negative definite. This matrix is non-negative definite if and only if the following three conditions are met:}\\

\textbf{A.3.1:} $\Sigma_D$ \textit{is non-negative definite};\\
\textbf{A.3.2:} $\Sigma_{DB}\in\textbf{range}\{\Sigma_D\}$;\\
\textbf{A.3.3:} $\Sigma_B-\Sigma_{BD}{\Sigma_{D}^{\dagger}}\Sigma_{DB}$ \textit{is non-negative definite}.\\

\textbf{A.4} \citep{BL} \textit{If $C$ is any matrix all of whose columns are in $\textbf{range}\{A\}$, then $AA^{\dagger}C=C$. Conversely, if we have $AA^{\dagger}C=C$ then $C$ is in the $\textbf{range}\{A\}$.}\\ 

\textbf{A.5} \citep{BL} \textit{Consider a consistent system of linear equations $AX=B$, where $A\neq 0$ is a non-negative definite matrix of dimension $r$ and rank $r'$; B is some $r\times m$ matrix whose columns are contained in $\textbf{range}\{A\}$; and X is some $r\times m$ solution matrix. Then, $$X = A^{\dagger}B + (I-A^{\dagger}A)T$$ for an arbitrary $r\times m$ matrix $T$. All possible solutions can be generated by varying the arbitrary matrix $T$.}\\

\section*{Appendix B}

\subsection*{B.1 - Proof of Lemma 3.1}

$P_{\bz}(X) = P_{P_{\bz}(\bm)}(X)=\bA\;P_{\bz}(\bm) + \pr{\bU}$\\

\textit{Proof.}\\

We begin by showing that $P_{P_{\bz}(\bm)}(X)=\bA\;P_{\bz}(\bm) + \pr{\bU}$: 

\begin{align}
P_{\newG}(X) &= 
\pr{X} + \Cov(X,\newG)\Var(\newG)^{\dagger}(\newG-\pr{\newG})\\
&= \pr{\am+\bU} + \Cov(\am+\bU,\newG)\Var(\newG)^{\dagger}(\newG-\pr{\newG})\\
&= \bA\;\pr{\bm} + \pr{\bU} + \bA\;\Cov(\bm,\newG)\Var(\newG)^{\dagger}(\newG-\pr{\newG})\label{covvar1}\\
&= \bA\;\pr{\bm} + \pr{\bU} +\bA\;\Var(\newG)\Var(\newG)^{\dagger}(\newG-\pr{\newG})\label{covvar2}\\
&= \bA\;\pr{\bm} + \pr{\bU} +\bA\;(\newG-\pr{\newG})\label{varcancelled}\\
&= \bA\;\pr{\bm} + \pr{\bU} + \bA\;\newG - \bA\;\pr{\newG}\\
&= \bA\;\pr{\bm} + \pr{\bU} + \bA\;\newG - \bA\;\pr{\bm}\\
&= \bA\;\newG + \pr{\bU}
\end{align}

Note that \eqref{covvar1} follows from \eqref{covvar2} as $\Cov(\bm,\newG)=\Cov(\newG,\newG)=\Var(\newG)$. 
\begin{align}
\Cov(\mu_i,P_{\bz}(\mu_j))&=
\Cov\bigg(\mu_i,\Cov(\mu_i,\bz)\Var(\bz)^{\dagger}\bz\bigg)\\
&=\Cov(\mu_i,\bz)\Var(\bz)^{\dagger}\Cov(\bz,\mu_j)\label{same1}
\end{align}

\begin{align}
\Cov(P_{\bz}(\mu_i),P_{\bz}(\mu_j))&=
\Cov\bigg(\Cov(\mu_i,\bz)\Var(\bz)^{\dagger}\bz\;,\;\Cov(\mu_j,\bz)\Var(\bz)^{\dagger}\bz\bigg)\\
&=\Cov(\mu_i,\bz)\Var(\bz)^{\dagger}\Cov(\bz,\bz)\Var(\bz)^{\dagger}\Cov(\bz,\mu_j)\\
&=\Cov(\mu_i,\bz)\Var(\bz)^{\dagger}\Var(\bz)\Var(\bz)^{\dagger}\Cov(\bz,\mu_j)\label{moore1}\\
&=\Cov(\mu_i,\bz)\Var(\bz)^{\dagger}\Cov(\bz,\mu_j)\label{moore2}
\end{align}

\eqref{moore1}-\eqref{moore2} holds as $\Var(\bz)^{\dagger}\Var(\bz)\Var(\bz)^{\dagger} = \Var(\bz)^{\dagger}$ by the definition of the Moore-Penrose generalised inverse (see Section 1.1 of \cite{GIB}). From \eqref{same1} and \eqref{moore2} we can see that the block elements of $\Cov(\bm,\newG)$ and $\Var(\newG)$ are indeed identical.\\

If $\Var(\newG)$ is of full rank, \eqref{covvar2}-\eqref{varcancelled} holds as $\Var(\newG)\Var(\newG)^{\dagger}=I$ by \textbf{A.2}. Alternatively, $$\Var(\newG)\Var(\newG)^{\dagger}(\newG-\pr{\newG}) = (\newG-\pr{\newG})$$ by \textbf{A.4}.\\

We now show that $\bA\;\newG + \pr{\bU}=P_{\bz}(X)$. 

\begin{equation*}
    \begin{split}
    \bA\;\newG + \pr{\bU} &=
    P_{\bz}(\bA\bm) + P(\bU) \\
    &= P_{\bz}(\bA\bm+\bU)\\
    &= P_{\bz}(X)\\
    \end{split}
\end{equation*}

Lemma \textcolor{red}{3.1} follows. 

\qed

\subsection*{B.2 - Proof of Corollary 3.1}

If we denote the sample means by $\overline{\bz} = \left[\;\overline{Z}_1,...,\overline{Z}_m\;\right]$ where $$\overline{Z}_i = \frac{1}{n_i}\sum_{j=1}^{n_i}Z_{ij}$$ it is sufficient to utilise the sample means from each co-exchangeable class $\overline{\bz}$ to update $\bm$ (i.e. $\lfloor \bz \indep \bm \rfloor \;/\; \overline{\bz}$).\\

\textit{Proof.}\\

$\lfloor \; Z_{ij} \indep \mu_i \rfloor \;/\; \overline{Z}_i$ (see equation 6.51 in \citep{BL}). This means that an individual member of a class $i$ is orthogonal to the mean of class $i$ given the sample mean of class $i$ for any $i$.\\

It remains to show that an individual member of class $i$ is orthogonal to the mean of class $k$ given the sample mean of class $i$ for any $i\neq k$. By \textbf{A.1}, it is sufficient to show that $\Cov(Z_{ij},\mu_k) = \Cov(Z_{ij},\overline{Z}_i)\Var(\overline{Z}_i)^{\dagger}\Cov(\overline{Z}_i,\mu_k)$:

\begin{align}
\Cov(Z_{ij},\overline{Z}_i)\Var(\overline{Z}_i)^{\dagger}\Cov(\overline{Z}_i,\mu_k)&=\Cov(\overline{Z}_i,\overline{Z}_i)\Var(\overline{Z}_i)^{\dagger}\Cov(\overline{Z}_i,\mu_k)\label{6.10}\\
    &=\Var(\overline{Z}_i)\Var(\overline{Z}_i)^{\dagger}\Cov(\overline{Z}_i,\mu_k)\label{16}\\
    &=\Cov(\overline{Z}_i,\mu_k)\label{17}\\
    &=\Cov(Z_{ij},\mu_k)\label{18}
\end{align}

\eqref{6.10} relies on the fact that $\Cov(Z_{ij},\overline{Z}_i) = \frac{1}{n_i}\sum_{j=1}^{n_i}\Cov(Z_{ij},\overline{Z}_i) = \Cov(\overline{Z}_i,\overline{Z}_i)$ (see Section 6.10 in \citep{BL}). \eqref{16}-\eqref{17} holds by \textbf{A.2} if $\Var(\overline{Z}_i)$ is full rank or, alternatively, by \textbf{A.3.2} in conjunction with \textbf{A.4}. \eqref{17}-\eqref{18} holds as both \eqref{17} and \eqref{18} are equal to $\Cov(\mu_i,\mu_k)$.\\
 
Therefore, $\lfloor Z_{ij} \indep \mu_k \rfloor / \overline{Z}_i.$\\

As $\lfloor \; Z_{ij} \indep \mu_i \rfloor \;/\; \overline{Z}_i$ and $\lfloor Z_{ij} \indep \mu_k \rfloor \;/\; \overline{Z}_i$, it follows that $\lfloor \bz \indep \bm \rfloor \;/\; \overline{\bz}$.

\qed \\

\subsection*{B.3 - Proof of Lemma 3.2}
 
$ P_{\mathcal{Z}}(X) \rightarrow \am + P(\bU)$ as $n_i \rightarrow \infty$ $\forall i$.\\

\textit{Proof.}\\

Recall that $\mathcal{Z}=P_{\overline{\bz}}(\bm)$. When $n_i \rightarrow \infty$, $\overline{Z}_i \rightarrow \mu_i$ $\forall i$ (see Section 6.9 of \citep{BL}). As $ \overline{Z}_i \rightarrow \mu_i $, $P_{\overline{Z}_i}(\mu_i) \rightarrow \mu_i$ $\forall i$. Equivalently, $P_{\overline{\bz}}(\bm) \rightarrow \bm$. Therefore, $P_{\mathcal{Z}}(X) \rightarrow \am + P(\bU)$ as $n_i \rightarrow \infty$ $\forall i$. \\

\qed

\section*{Appendix C}

\begin{table}[hbt!]
\begin{center}
\begin{tabular}{ c|c|c|c|c|c|c|c|c } 
         
            &
         \textbf{L=500} &
         \textbf{SE=238} & 
         \textbf{NW=210} & 
         \textbf{E=123} & 
         \textbf{EM=173} &
         \textbf{WM=142}  &
         \textbf{Y=170}  &
         \textbf{NE=170}  \\
\hline
\textbf{L} & $\times$ & $\times$ & $\times$ & $\times$ & $\times$ & $\times$ & $\times$ & $\times$   \\ 
\hline 
\textbf{SE} & 200 & $\times$ & $\times$ & $\times$ & $\times$ & $\times$ & $\times$ & $\times$   \\ 
\hline 
\textbf{NW} & 180 & 190 & $\times$ & $\times$ & $\times$ & $\times$ & $\times$ & $\times$   \\ 
\hline 
\textbf{E} & 85 & 100 & 105 & $\times$ & $\times$ & $\times$ & $\times$ & $\times$   \\ 
\hline 
\textbf{EM} & 95 & 115 & 120 & 140 & $\times$ & $\times$ & $\times$ & $\times$  \\ 
\hline 
\textbf{WM} & 50 & 55 & 75 & 80 & 110 & $\times$ & $\times$ & $\times$   \\ 
\hline 
\textbf{Y} & 50 & 55 & 75 & 80 & 105 & 130 & $\times$ & $\times$   \\ 
\hline 
\textbf{NE} & 50 & 55 & 70 & 75 & 85 & 95 & 130 & $\times$   \\ 
\hline
\textbf{SW} & 25 & 27 & 28 & 28 & 29 & 40 & 42 & 43 \\

\end{tabular}
\end{center}
\caption{Elicitation for  $\Var(X)$.}
\label{VYE}
\end{table}

The method of \cite{kadane1980interactive} allows one to build up a covariance matrix iteratively through the use of \eqref{iter}, where $g_{i+1}$ is found via the solving of a system of linear equations (equation 3.10 in \cite{kadane1980interactive}). The parallels needed to apply this method within this context are intuitive given that the expressions used for the mean and variance of the multivariate t-distribution in the application of \cite{kadane1980interactive} aligns with the form of the adjusted expectation and variance in a Bayes Linear analysis. This system is derived from a specific collection of conditional previsions, the collection we elicited from our expert shown in Table \ref{VYE}. The crux of this method is that it forms a correlation matrix through considering how these conditional previsions evolve given new hypothetical values. 

\begin{align}\label{iter}
    \text{U}_{i+1} &= \begin{bmatrix}
    \text{U}_i & \text{U}_i\text{g}_{i+1}\\
    \text{g}_{i+1}^{'}\text{U}_i & \Var(X_{\text{i}+1})\\
    \end{bmatrix}
\end{align}

We'll outline the first iteration of this method fully (i.e. the step in which $i=1$). Firstly, we make our prior specifications on $X_{L}$, $P(X_{\text{L}})=400$ and $\Var(X_{\text{L}})=40000$ (this is taken to be $U_1$ in \eqref{iter}). We then select the initial value, $N_L$, of cases in London that is going to be used to elicit the previsions regarding the South East. As explained in the main text, this is taken to be $N_L=400+\frac{1}{2}\sqrt{40000}=500$. We then elicit the conditional previsions $P(X_{\text{SE}}|N_L=500)=200$ and $\Var(X_{\text{SE}}|N_L=500)=5625$. These are all the specifications required to solve for $g_2=0.2$, via equation 3.10a in \cite{kadane1980interactive}. 

Then, by equation 3.11 in \cite{kadane1980interactive}, $\Var(X_{SE})=\Var(X_{SE}|N_L)+g_2 U_1 g_2=7225$. This form of the prior variance on the South East ensures that the matrix $U_2$ is positive definite (see equation 3.12 and associated explanation in \cite{kadane1980interactive}).  

\begin{align}
    \text{U}_{2} &= \begin{bmatrix}
    40000 & 8000\\
    8000 & 7225\\
    \end{bmatrix}
\end{align}

The top row of Table \ref{VYE} shows the initial values assigned to each region for the conditional prevision assessments. These specifications build up as you move across the columns. For example, the first column contains the expected number of S$-$ cases in the remaining regions given that there are 500 cases in London. The second column then contains the expected number of S$-$ cases in the remaining regions given that there are 500 cases in London and 238 cases in the South East, etc.

\cite{al1998elicitation} found that the conditional variances of the random variables which appeared later in the ordering tended to be overinflated and proposed rescaling the underlying correlation matrix by the marginal variances instead. In our case, this was fundamental to the use of the method as the marginal variances were elicited given the knowledge that our quantity of interest was positive. If we were to use the matrix elicited via \cite{kadane1980interactive} without rescaling, our prior $\pm2$ standard deviation range on the respective regions would, in some cases, encompass negative values (and thus imply that our expert believed this to be a feasible occurrence).

\begin{table}[hbt!]
\begin{center}
\begin{tabular}{ c|c|c|c|c|c|c|c|c|c } 
         
            &
         \textbf{L} &
         \textbf{SE} & 
         \textbf{NW} & 
         \textbf{E} & 
         \textbf{EM} &
         \textbf{WM}  &
         \textbf{Y}  &
         \textbf{NE} &
         \textbf{SW} \\
\hline
\textbf{L} & 1 & 0.47 & 0.80 & 0.21 & 0.33 & 0.20 & 0.17 & 0.17 & 0.18  \\ 
\hline 
\textbf{SE} & 0.47 & 1 & 0.61 & 0.65 & 0.54 & 0.18 & 0.16 & 0.16 & 0.15  \\ 
\hline 
\textbf{NW} & 0.80 & 0.61 & 1 & 0.44 & 0.44 & 0.41 & 0.34 & 0.30 & 0.18  \\ 
\hline 
\textbf{E} & 0.21 & 0.65 & 0.44 & 1 & 0.68 & 0.26 & 0.22 & 0.20 & 0.09  \\ 
\hline 
\textbf{EM} & 0.33 & 0.54 & 0.44 & 0.68 & 1 & 0.63 & 0.47 & 0.28 & 0.12  \\ 
\hline 
\textbf{WM} & 0.20 & 0.18 & 0.41 & 0.26 & 0.63 & 1 & 0.72 & 0.37 & 0.33  \\ 
\hline 
\textbf{Y} & 0.17 & 0.16 & 0.34 & 0.22 & 0.47 & 0.72 & 1 & 0.69 & 0.28  \\ 
\hline 
\textbf{NE} & 0.17 & 0.16 & 0.30 & 0.20 & 0.28 & 0.37 & 0.69 & 1 & 0.19  \\ 
\hline 
\textbf{SW} & 0.18 & 0.15 & 0.18 & 0.09 & 0.12 & 0.33 & 0.28 & 0.19 & 1  \\ 
\end{tabular}
\end{center}
\caption{Correlation matrix for $X$.}
\end{table}

\newpage

\section*{Appendix D}
For completeness we show the derived matrices for $\bA$ and the prevision of $\bU$.
\subsection*{D.1 - $\bA$}

\begin{equation*}
\bA_1 = 
\begin{bmatrix}
0.20 & 0.00 & 0.00 & 0.00 & 0.00 & 0.00 & 0.00 & 0.00 & 0.00\\
-0.70 & 0.34 & 2.24 & -1.68 & 1.75 & -2.87 & -0.25 & -0.13  & 1.29\\
-0.23 & 0.18 & 0.89 & -0.71 & 0.47 & -0.75 & -0.08 & -0.02 & 0.31\\
0.07 & 0.04 & -0.29 & 0.55 & -0.20 & 0.38 & 0.01 & 0.05 & -0.14\\
-0.01 & 0.03 & -0.02 & -0.02 &  0.38 &  0.04 & -0.01 &  0.03 & 0.01\\
0.15 & 0.06 & -0.55 & 0.41 & -0.57 &  1.50 & 0.04 & 0.07 & -0.33\\
0.11 & 0.04 & -0.41 &  0.31 & -0.43 &  1.05 & 0.08 & 0.14 & -0.31\\
0.06 & 0.03 & -0.24 & 0.14 & -0.21 &  0.69 & -0.57 & 0.92 & -0.12\\
-0.02 & 0.00 & 0.05 & -0.05 & 0.05 & -0.02 & -0.13 & 0.06 &  0.78
\end{bmatrix}
\end{equation*}

\begin{equation*}
\bA_2 = 
\begin{bmatrix}
0.49 & 0.00 & 0.00 & 0.00 & 0.00 & 0.00 & 0.00 & 0.00 & 0.00\\
 0.58 & 0.15 & -1.69 & 1.27 & -1.34 & 2.22 & 0.19 & 0.10 & -1.21\\
 0.15 & -0.41 & 0.20 & 0.63 & -0.06 & 0.05 & 0.03 & -0.04 & 0.07\\
-0.04 & -0.19 & 0.24 &  0.37 & 0.22 & -0.43 & -0.01 & -0.05 & 0.23\\
 0.01 & -0.16 & 0.09 & 0.11 & 0.44 & -0.20 & 0.00 & -0.04 & 0.12\\
-0.06 & -0.04 & 0.24 & -0.14 & 0.20 & -0.02 & -0.02 & -0.03 & 0.12\\
-0.05 & -0.03 & 0.18 & -0.11 & 0.15 & -0.29 & 0.37 & -0.03 & 0.08\\
-0.02 & -0.03 & 0.11 & -0.05 & 0.09 & -0.19 & 0.05 & 0.28 & 0.06\\
 0.01 & -0.01 & -0.03 & 0.04 & -0.04 & 0.07 & 0.02 & 0.00 & 0.10
\end{bmatrix}
\end{equation*}

\begin{equation*}
\bA_3 = 
\begin{bmatrix}
0.47 & 0.00 & 0.00 & 0.00 & 0.00 & 0.00 & 0.00 & 0.00 & 0.00\\
 0.19 & 0.57 & -0.72 &  0.54 & -0.55 &  0.89 &  0.08 &  0.04 & -0.27\\
 0.15 & 0.18 & -0.14 &  0.24 & -0.50 & 0.85 & 0.06 & 0.05 & -0.51\\
-0.03 & 0.14 & 0.05 & 0.15 & -0.03 & 0.04 &  0.00 & 0.00 & -0.10\\
 0.02 & 0.12 & -0.08 & -0.06 & 0.22 &  0.18 & 0.02 & 0.00 & -0.16\\
-0.09 & -0.03 & 0.32 & -0.28 &  0.41 & -0.54 & -0.02 & -0.05  & 0.24\\
-0.06 & -0.02 & 0.24 & -0.21 &  0.31 & -0.88 & 0.68 & -0.14 & 0.27\\
-0.03 & -0.01 & 0.14 & -0.09 & 0.14 & -0.59 & 0.64 & -0.26 & 0.06\\
 0.01 & 0.01 & -0.02 & 0.01 & -0.01 & -0.07 & 0.13 & -0.07 & 0.13
\end{bmatrix}
\end{equation*}

\subsection*{D.2 - $P(\bU)$}

\begin{equation*}
    P(\bU) = \begin{bmatrix}
    -62.64\\
    -19.15\\
    -21.48\\
    -5.12\\
    -6.29\\
    -0.62\\
    -1.62\\
    -0.80\\
    -0.60\\
    \end{bmatrix}
\end{equation*}

\section*{Additional Prior Quantities}

The full form of all other prior specifications utilised in the main text can be retrieved via running the provided code. If not fully printed in the main text or here, this is due either to the fact that a) the quantity is straightforwardly derivable from a quantity that has already been stated (e.g. we defer $\Var(\mu_i)$ for $i=1,2,3$ to code as it is obtained by scaling $\Var(X)$ which is included in its entirety in the main text) or b) the elements of a matrix we deem to be most significant are stated and discussed alongside explanation as to how to derive the off-diagonal elements (e.g. we include the diagonal elements of $\Var_{\mathcal{Z}}(X)$ and $\Var(\bU)$ in Table 5 of the main text to enable us to discuss proportions of uncertainty resolved region-by-region and defer the off-diagonal elements to the code). 

\bibliographystyle{agsm}
\bibliography{sample} 

\end{document}